\pdfoutput=1

\documentclass[11pt]{article}

\usepackage{naacl2021}

\usepackage{times}
\usepackage{latexsym}

\usepackage[T1]{fontenc}

\usepackage[utf8]{inputenc}

\usepackage{microtype}

\usepackage{url}
\usepackage{latexsym}
\usepackage{xspace}
\usepackage{hyperref}
\usepackage{graphicx}
\usepackage{multirow}
\usepackage{amsmath}

\usepackage{booktabs}
\usepackage[inline]{enumitem}
\usepackage[caption=false]{subfig}

\newcommand\F{\textit{{$\text{F}_1$}}\xspace}
\newcommand\Fa{\textit{{$\text{F}_1$}-a}\xspace}
\newcommand\Fi{\textit{{$\text{F}_1$}-i}\xspace}
\newcommand\Fs{\textit{{$\text{F}_1$}-s}\xspace}

\newcommand\accs{\textit{acc-s}\xspace}

\newcommand{\eg}{\textit{e.\,g.}\xspace}

\usepackage{annotates}

\usepackage{array}
\newcolumntype{P}[1]{>{\centering\arraybackslash}p{#1}}
\newcommand{\sd}[1]{\par \scriptsize #1}

\usepackage{xcolor}
\definecolor{lightblue}{RGB}{200,245,255}
\definecolor{lightred}{RGB}{255,200,200}
\definecolor{lightgrey}{RGB}{215,215,215}
\definecolor{green}{RGB}{155,211,203}
\definecolor{pink}{RGB}{243,156,164}

\newcommand{\posbox}[1]{{\setlength{\fboxsep}{1pt}\colorbox{lightblue}{#1}}}
\newcommand{\negbox}[1]{{\setlength{\fboxsep}{1pt}\colorbox{lightred}{#1}}}
\newcommand{\neubox}[1]{{\setlength{\fboxsep}{1pt}\colorbox{lightgrey}{#1}}}

\newcommand{\pinkbox}[1]{{\setlength{\fboxsep}{1pt}\colorbox{pink}{#1}}}
\newcommand{\greenbox}[1]{{\setlength{\fboxsep}{1pt}\colorbox{green}{#1}}}

\usepackage[page, title]{appendix}

\title{Multi-task Learning of Negation and Speculation\\ for Targeted Sentiment Classification} 

\author{Andrew Moore\thanks{~~The authors contributed equally.} \\
School of Computing \\and Communications,\\
  Lancaster University \\
  {\tt a.moore@lancaster.ac.uk} \\\And
  Jeremy Barnes \\
  University of Oslo\\
  Department of Informatics\\
 {\tt jeremycb@ifi.uio.no}}

\begin{document}
\maketitle
\begin{abstract}
The majority of work in targeted sentiment analysis has concentrated on finding better methods to improve the overall results. Within this paper we show that these models are not robust to linguistic phenomena, specifically negation and speculation. In this paper, we propose a multi-task learning method to incorporate information from syntactic and semantic auxiliary tasks, including negation and speculation scope detection, to create English-language models that are more robust to these phenomena. Further we create two challenge datasets to evaluate model performance on negated and speculative samples. We find that multi-task models and transfer learning via language modelling can improve performance on these challenge datasets, but the overall performances indicate that there is still much room for improvement. We release both the datasets and the source code at \url{https://github.com/jerbarnes/multitask_negation_for_targeted_sentiment}.
\end{abstract}

\section{Introduction}
\label{sec:intro}

Targeted sentiment analysis (TSA) involves jointly predicting entities which are the targets of an opinion, as well as the polarity expressed towards them \cite{mitchell-etal-2013-open}. The TSA task, which is part of the larger set of fine-grained sentiment analysis tasks, can enable companies to provide better recommendations \cite{bauman2017aspect}, as well as give digital humanities scholars a quantitative approach to identifying how sentiment and emotions develop in literature \cite{Alm2005,kim-klinger-2019-frowning}.

Modelling TSA has moved from sequence labeling using conditional random fields (CRFs) \cite{mitchell-etal-2013-open} or Recurrent Neural Networks (RNN) \cite{Zhang2015,katiyar-cardie-2016-investigating,ma-etal-2018-joint}, to Transformer models \cite{hu-etal-2019-open}. However, all these improvements have concentrated on making the best of the relatively small task-specific datasets. As annotation for fine-grained sentiment is difficult and often has low inter-annotator agreement \cite{Wiebe2005b,oevrelid2020-finegrained}, this data tends to be small and of varying quality.
This lack of high-quality training data prevents TSA models from learning complex, compositional linguistic phenomena. For sentence-level sentiment classification, incorporating compositional information from relatively small amounts of negation or speculation data improves both robustness and general performance \cite{councill-etal-2010-great,Cruz2016,barnes-etal-2020}. 
Furthermore, transfer learning via language-modelling also improves fine-grained sentiment analysis \cite{hu-etal-2019-open,li-etal-2019-exploiting}. In this paper, we wish to explore \textbf{two research questions}:

\begin{enumerate}
    \item Does multi-task learning of negation and speculation lead to more robust targeted sentiment models?
    \item Does transfer learning based on language-modelling already incorporate this information in a way that is useful for targeted sentiment models?
\end{enumerate}

We explore a \textbf{multi-task learning} (\textbf{MTL}) approach to incorporate auxiliary task information in targeted sentiment classifiers in English
in order to investigate the effects of negation and speculation in detail, we also annotate two new challenge datasets which contain negated and speculative examples. 
We find that the performance is negatively affected by negation and speculation, but MTL and \textbf{transfer learning} (\textbf{TL}) models are more robust than \textbf{single task learning} (\textbf{STL}).
TL reduces the improvements of MTL, suggesting that TL is similarly effective at learning negation and speculation. The overall performance on the challenge datasets, however, confirms that there is still room for improvement.


The contributions of the paper are the following: 
\begin{enumerate*}[label=\roman*)]
    \item we introduce two English challenge datasets annotated for negation and speculation,
    \item we propose a multi-task model to incorporate negation and speculation information and evaluate it across four English datasets,
    \item Finally, using the challenge datasets, we show the quantitative effect of negation and speculation on TSA.
\end{enumerate*}

\section{Background and related work}
\label{sec:background}

\paragraph{Fine-grained sentiment analysis} is a complex task which can be broken into four subtasks \cite{Liu:15}: 
\begin{enumerate*}[label=\roman*)]
  \item opinion holder extraction,
  \item opinion target extraction,
  \item opinion expression extraction,
  \item and resolving the polarity relationship between the holder, target, and expression.
\end{enumerate*}
 From these four subtasks, targeted sentiment analysis (TSA) \cite{jin-etal-2009-lexicalized,chen-etal-2012-extracting,mitchell-etal-2013-open} reduces the fine-grained task to only the second and final subtasks, namely extracting the opinion target and the polarity towards it. 

English TSA datasets include MPQA \cite{Wiebe2005b}, the SemEval Laptop and Restaurant reviews \cite{pontiki-etal-2014-semeval,pontiki-etal-2016-semeval}, and Twitter datasets \cite{mitchell-etal-2013-open,wang-etal-2017-tdparse}. Further annotation projects have led to review datasets for Arabic, Dutch, French, Russian, and Spanish \cite{pontiki-etal-2016-semeval} and Twitter datasets for Spanish \cite{mitchell-etal-2013-open} and Turkish \cite{pontiki-etal-2016-semeval}. Prior work has also explored the effects of different phenomena on TSA through error analysis and challenge datasets. \newcite{wang-etal-2017-tdparse}, \newcite{xue-li-2018-aspect}, and \newcite{jiang-etal-2019-challenge} showed the difficulties of polarity classification of targets on texts with multiple different polarities through the distinct sentiment error splits, the hard split, and the MAMS challenge dataset respectively.  Both \newcite{Kaushik2020Learning} and \newcite{gardner2020evaluating} augment document sentiment datasets by asking annotators to create counterfactual examples for the IMDB dataset. 
More recently, \newcite{ribeiro-etal-2020-beyond} showed how sentence-level sentiment models are affected by various linguistic phenomena including negation, semantic role labelling, temporal changes, and name entity recognition. 
Previous approaches to modelling TSA have often relied on general sequence labelling models, \eg CRFs \cite{mitchell-etal-2013-open}, probabilistic graphical models \cite{Klinger2013}, RNNs \cite{zhang-etal-2015-neural,ma-etal-2018-joint}, and more recently pretrained Transformer models \cite{li-etal-2019-exploiting}. 

\paragraph{Multi-task and transfer learning}
\label{sec:related:mtl}

The main idea of MTL \cite{Caruana93multitasklearning} is that a model which receives signal from two or more correlated tasks will more quickly develop a useful inductive bias, allowing it to generalize better. This approach has gained traction in NLP, where several benchmark datasets have been created \cite{wang2018glue,wang2019superglue}. Under some circumstances, MTL can also be seen as a kind of data augmentation, where a model takes advantage of extra training data available in an auxiliary task to improve the main task \cite{kshirsagar-etal-2015-frame,plank-2016-keystroke}. 
Much of MTL uses \textit{hard parameter sharing} \cite{Caruana93multitasklearning}, which shares all parameters across some layers of a neural network. When the main task and auxiliary task are closely related, this approach can be an effective way to improve model performance \cite{Collobert2011a,peng-dredze-2017-multi,Alonso2017,Augenstein2018}, although it is often preferable to make predictions for low-level auxiliary tasks at lower layers of a multi-layer MTL setup \cite{Sogaard2016}, which we refer to as \emph{hierarchical MTL}.


Transfer learning methods \cite{Mikolov2013word2vec,peters-etal-2018-deep,devlin-etal-2019-bert} can leverage unlabeled data, but require training large models on large amounts of data. However, it seems even these models can be sensitive to negation \cite{ettinger-2020-bert,ribeiro-etal-2020-beyond,kassner-schutze-2020-negated}

Specific to TSA, previous research has used MTL to incorporate document-level sentiment \cite{he-etal-2019-interactive}, or to jointly learn to extract opinion expressions \cite{li-etal-2019-exploiting,chen-qian-2020-relation}.

\paragraph{Negation and Speculation Detection}

As negation is such a common linguistic phenomenon and one that has a direct impact on sentiment, previous work has shown that incorporating negation information is crucial for accurate sentiment prediction. Feature-based approaches did this by including features from negation detection modules \cite{Das2007,councill-etal-2010-great,Lapponi2012}, while it has now become more common to assume that neural models learn negation features in an end-to-end fashion \cite{Socher2013b}. However, recent research suggests that end-to-end models are not able to robustly interpret the effect of negation on sentiment \cite{barnes-etal-2019}, and that explicitly learning negation can improve sentiment results \cite{Barnes2019-neges,barnes-etal-2020}.

On the other hand, speculation refers to whether a statement is described as a fact, a possibility, or a counterfact \cite{sauri2009}. Although there are fewer speculation annotated corpora available \cite{Vin:Sza:Far:08,kim-etal-2013-genia,Konstantinova2012}, including speculation information has shown promise for improving sentiment analysis at document-level \cite{Cruz2016}.

There has, however, been little research on how these phenomena specifically affect fine-grained approaches to sentiment analysis. This is important because, compared to document- or sentence-level tasks where there is often a certain redundancy in sentiment signal, for fine-grained tasks negation and speculation often completely change the sentiment (see Table \ref{tab:labelflip}), making their identification and integration within a fine-grained sentiment models essential to resolve.

\section{Data}

We perform the main experiments on four English language datasets: The \textbf{Laptop} dataset from SemEval 2014 \cite{pontiki-etal-2014-semeval}, the \textbf{Restaurant} dataset which combines the SemEval 2014 \cite{pontiki-etal-2014-semeval}, 2015 \cite{pontiki-etal-2015-semeval}, and 2016 \cite{pontiki-etal-2016-semeval}, the Multi-aspect Multi-sentiment (\textbf{MAMS}) dataset \cite{jiang-etal-2019-challenge}, and finally the Multi-perspective Question Answering (\textbf{MPQA}) dataset \cite{Wiebe2005b}\footnote{All datasets contain the following three sentiment classes positive, neutral, and negative. The MPQA dataset also includes a fourth rare class, both. Table \ref{table:appendix_label_distribution_stats} of Appendix \ref{appendix_label_distribution}.} shows the distribution of the sentiment classes . We take the pre-processed Laptop and Restaurant datasets from \newcite{li2019unified}, and use the train, dev, and test splits that they provide. 
We use the NLTK word tokenizer to tokenise the Laptop, Restaurant, and MPQA datasets and Spacy for the MAMS dataset.

We choose datasets that differ largely in their domain, size, and annotation style in order to determine if any trends we see are robust to these data characteristics or whether they are instead correlated. We convert all datasets to a targeted setup by extracting only the aspect targets and their polarity. We use the unified tagging scheme\footnote{This is also known as collapsed tagging scheme \cite{hu-etal-2019-open}} following recent work \cite{li2019unified,li-etal-2019-exploiting} and convert all data to BIOUL format\footnote{BIOUL format tags each token as either \textbf{B}: beginning token, \textbf{I}: inside token, \textbf{O}: outside token, \textbf{U}: unit (single token), or \textbf{L}: last token.} 
with unified sentiment tags, \eg \emph{B-POS} for a beginning tag with a positive sentiment, so that we can cast the TSA problem as a sequence labeling task.

The statistics for these datasets are shown in Table \ref{tab:datastats}. MAMS has the largest number of training targets (11,162), followed by Restaurant (3,896), Laptop (2,044) and finally MPQA has the fewest (1,264). MPQA, however, has the longest average targets (6.3 tokens) compared to 1.3-1.5 for the other datasets. This derives from the fact that entire phrases are often targets in MPQA. Finally, due to the annotation criteria, the MAMS data also has the highest number of sentences with multiple aspects with multiple polarities -- nearly 100\% in train, compared to less than 10\% for Restaurant.

\begin{table*}[]
    \centering
    \resizebox{\textwidth}{!}{%
    \begin{tabular}{lrrrrrrrrrrrrr}
    \toprule
          &  \multicolumn{4}{c}{\textbf{Train}} & \multicolumn{4}{c}{\textbf{Dev}} & \multicolumn{4}{c}{\textbf{Test}} \\
          \cmidrule(lr){2-5}\cmidrule(lr){6-9}\cmidrule(lr){10-14}
     & sents. & targs. & len. & mult. & sents. & targs. & len. & mult. & sents. & targs. & len. & mult. & IAA\\
    \cmidrule(lr){2-2}\cmidrule(lr){3-3}\cmidrule(lr){4-4}\cmidrule(lr){5-5}\cmidrule(lr){6-6}\cmidrule(lr){7-7}\cmidrule(lr){8-8}\cmidrule(lr){9-9}\cmidrule(lr){10-10}\cmidrule(lr){11-11}\cmidrule(lr){12-12}\cmidrule(lr){13-13}\cmidrule(lr){14-14}
    Laptop     & 2,741  &2,044 & 1.5 & 136 & 304 &256 & 1.5 & 18 & 800 &634 & 1.6 & 38 & 0.67\\
    Laptop$_{Neg}$ & -- &  -- & -- & -- & 147 & 181 & 1.5 & 41 & 403 &470 & 1.6 & 79 & 0.70\\
    Laptop$_{Spec}$ & -- &-- & -- & -- & 110 &142 & 1.4 & 10 & 208 & 220 & 1.5 & 19 & 0.64\\ 
    Restaurant & 3,490 & 3,896 & 1.4 & 312 & 387 &414 & 1.4 & 34 & 2,158 &2,288 & 1.4 & 136 & 0.71 \\
    Restaurant$_{Neg}$ & -- & -- & -- & -- & 198 & 274 & 1.4 & 61 & 818 & 1,013 & 1.4 & 161 & 0.66\\
    Restaurant$_{Spec}$ & -- & -- & -- & -- & 138 &200 & 1.3 & 35 & 400 & 451 & 1.4 & 49 & 0.66\\
    MAMS & 4,297 &11,162 & 1.3 & 4,287 & 500&1,329 & 1.3  & 498 &500 & 1,332 & 1.3& 500 & -\\
    MPQA & 4,195 & 1,264 & 6.3 & 94 & 1,389&400 & 5.4 & 29 & 1,620 & 365 & 6.7 & 22 & -\\
    \bottomrule
    \end{tabular}
    }
    \caption{Statistics for the sentiment datasets used in the experiments. The table indicates the number of sentences in each split (sents.),  the number of targets (targs.), the average length of the targets (len.), as well as how many sentences in each have multiple targets with differing polarity (mult.). IAA scores are reported on a subset of the data.}
    \label{tab:datastats}
\end{table*}

\subsection{Annotation for negation and speculation}

Although negation and speculation are prevalent in the original data -- negation and speculation occur in 13-25\% and 9-20\% of the sentences, respectively -- it is difficult to pry apart improvement on the original data with improvement on these two phenomena. Therefore, we further annotate the dev and test set for the Laptop and Restaurant datasets\footnote{For clarification this is the SemEval 2014 Laptop dataset and the 2014, 2015, and 2016 combined Restaurant dataset.}, and when possible\footnote{While inserting negation into new sentences is quite trivial, as one can always negate full clauses, \eg It's good $\rightarrow$ It's not true that it's good, adding speculation often requires rewording of the sentence. We did not include sentences that speculation made unnatural.}, insert negation and speculation cues into sentences lacking them, which we call Laptop$_{Neg}$, Laptop$_{Spec}$, Restaurant$_{Neg}$, and Restaurant$_{Spec}$. Inserting negation and speculation cues often leads to a change in polarity from the original annotation, as shown in the example in Table \ref{tab:labelflip}. We finally keep all sentences that contain a negation or speculation cue, including those that occur naturally in the data. As this process could introduce errors regarding the polarity expressed towards the targets, we doubly annotate the polarity for 50 sentences from the original dev data, the negated dev data, and the speculation dev data and calculate Cohen's Kappa scores. The statistics and inter-annotator agreement scores (IAA) are shown in Table \ref{tab:datastats}\footnote{Table \ref{table:appendix_label_distribution_stats} of Appendix \ref{appendix_label_distribution} shows the distribution of the sentiment classes.}. The new annotations have similarly high IAA scores (0.66-0.70) to the original data (0.67-0.71), confirming the quality of the annotations.

\begin{table*}[!h]

 \centering
    \begin{tabular}{ll}
    \toprule
    original     & this is good, inexpensive \posbox{sushi}. \\
    negated     & this is \textbf{not} good, inexpensive \negbox{sushi}. \\
    speculative & \textbf{I'm not sure if} this is good, inexpensive \neubox{sushi}. \\
    \bottomrule
    \end{tabular}
    \caption{Example of how adding negation and speculation can change the polarity of a target (added tokens are shown in \textbf{bold}). While in the original, the target ``sushi'' has a \posbox{positive} polarity, in the negated example it is \negbox{negative}, and in the speculative example it is \neubox{neutral}.}
    \label{tab:labelflip}
\end{table*}

\subsection{Auxiliary task data}

For the multi-task learning experiments, we use six auxiliary tasks: negation scope detection using the Conan Doyle (\textbf{NEG$_{CD}$}) \cite{Mor:Dae:12}, both negation detection (\textbf{NEG$_{SFU}$}) and speculation detection (\textbf{SPEC}) on the SFU$_{NegSpec}$ dataset  \cite{Konstantinova2012}, and Universal Part-of-Speech tagging (\textbf{UPOS}), Dependency Relation prediction (\textbf{DR}) and prediction of full lexical analysis (\textbf{LEX}) on the Streusle dataset \cite{schneider-smith-2015-corpus}. We show the train, dev, test splits, as well as the number of labels, label entropy and label kurtosis \cite{Alonso2017} in Table \ref{tab:auxiliary_stats}. An example sentence with auxiliary labels is shown in Appendix \ref{appendix_auxiliary_tasks}. Although it may appear that the SFU dataset is an order of magnitude larger than the Conan Doyle dataset, in reality, most of the training sentences do not contain annotations, leaving similar sized data if these are filtered. Similar to the sentiment data, we convert the auxiliary tasks to BIO format and treat them as sequence labelling tasks.

\begin{table*}[]
    \centering
    \begin{tabular}{lrrrrrr}
    \toprule
    & train & dev & test & \# labels & label entropy & label kurtosis \\
    \cmidrule(lr){2-2}\cmidrule(lr){3-3}\cmidrule(lr){4-4}\cmidrule(lr){5-5}\cmidrule(lr){6-6}\cmidrule(lr){7-7}
    NEG$_{CD}$     & 842 & 144 & 235 & 5 & 1.0 & -0.8 \\
    NEG$_{SFU}$   &  13,712 & 1,713 & 1,703 & 5 & 0.2 & 0.2 \\
    SPEC  &  13,712 & 1,713 & 1,703 & 5 & 0.1 & 0.2 \\
    UPOS  &  2,723 & 554 & 535 & 17 & 2.5 & -0.6\\
    DR    &  2,723 & 554 & 535 & 49 & 3.1 & 1.3\\
    LEX   &  2,723 & 554 & 535 & 570 & 3.9 & 75.7\\
    \bottomrule
    \end{tabular}
    \caption{Statistics for the auxiliary datasets.}
    \label{tab:auxiliary_stats}
\end{table*}

\section{Experiments}

\begin{figure}[!h]
    \centering
    \includegraphics[scale=0.4]{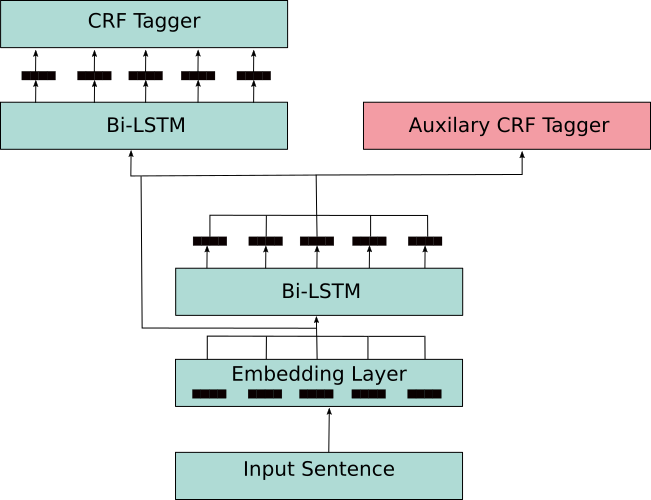}
    \caption{The overall architecture where the STL model contains all of the \greenbox{green layers} and the MTL uses the additional \pinkbox{pink auxiliary CRF tagger}. The second Bi-LSTM has a skip connection from the embedding layer which concatenates the word embeddings with the output from the first Bi-LSTM.}
    \label{fig:architecture}
\end{figure}

We experiment with a single task baseline (STL) and a hierarchical multi-task model with a skip-connection (MTL), both of which are shown in Figure \ref{fig:architecture}. For the STL model, we first embed a sentence and then pass the embeddings to a Bidirectional LSTM (Bi-LSTM). These features are then concatenated to the input embeddings and fed to the second Bi-LSTM layer, ending with the token-wise sentiment predictions from the CRF tagger. For the MTL model, we additionally use the output of the first Bi-LSTM layer as features for the separate auxiliary task CRF tagger. As seen from Figure \ref{fig:architecture}, the STL model and the MTL main task model use the same the green layers. The MTL additionally uses the pink layer for the auxiliary task, adding less than 3.4\% trainable parameters\footnote{The STL model had 1,785,967 parameters of which 364,042 were trainable as the embedding layer was frozen.} for all auxiliary tasks except LEX, which adds 221.4\% due to the large label set (see Table \ref{tab:auxiliary_stats}). Furthermore, at inference time the MTL model is as efficient as STL, given that it only uses the green layers when predicting the targeted sentiment, of which this is empirically shown in Table \ref{table:inference_times} of Appendix \ref{appendix:additional_reproducibility_statistics}.

\textbf{Embeddings:} For the embedding layer, we perform experiments using 300 dimensional \textbf{GloVe} embeddings \cite{Pennington2014}, as well as \textbf{TL} from Transformer ELMo embeddings \cite{peters-etal-2018-dissecting}\footnote{This is a 6 layer transformer model with a bi-directional language model objective that contains 56 million parameters excluding the softmax. In comparison BERT uses a masked language modelling objective and contains 110 and 340 million parameters for the base and large versions \cite{devlin-etal-2019-bert}.}. The GloVe embeddings are publicly available and trained on English Wikipedia and Gigaword data. For the MPQA dataset we use the Transformer ELMo from \newcite{peters-etal-2018-dissecting}\footnote{Found at \url{https://allennlp.org/elmo} under Transformer ELMo.} which was trained on the 1 billion word benchmark \cite{chelba2014one}. For the MAMS and Restaurant datasets we tuned a Transformer ELMo on 27 million (M) sentences from the 2019 Yelp review dataset\footnote{\url{https://www.yelp.com/dataset}}, and for the Laptop dataset on 28M sentences\footnote{More specifically there was 9M unique sentences and the model was trained for 3 epochs.} from the Amazon electronics reviews dataset \cite{mcauley2015image}\footnote{For full details of on how the fine tuned Transformer ELMo models were trained see \url{https://github.com/apmoore1/language-model}.}. Training these models on large amounts of in-domain data gives superior performance to models trained on more generic data, \eg BERT \cite{devlin-etal-2019-bert}. For all experiments we freeze the embedding layer in order to make the results between GloVe and TL more comparable with respect to the number of trainable parameters. For TL, 
we learn a summed weighting of all layers\footnote{For this Transformer ELMo it uses the output from the 6 transformer layers and the output from the non-contextualised character encoder, thus in total 7 layers are weighted and summed.}, as this is more effective than using the last layer \cite{peters-etal-2018-deep}. For more details on the number of parameters used for each model see Table \ref{table:parameter_table} in Appendix \ref{appendix:additional_reproducibility_statistics}.

\begin{table*}[!h]
    \centering
    \resizebox{\textwidth}{!}{
    \begin{tabular}{llP{1.2cm}P{1.2cm}P{1.2cm}P{1.2cm}P{1.2cm}P{1.2cm}P{1.2cm}P{1.2cm}}
\toprule
& & \multicolumn{2}{c}{Laptop} & \multicolumn{2}{c}{MAMS} & \multicolumn{2}{c}{Restaurant} & \multicolumn{2}{c}{MPQA} \\
\cmidrule(lr){2-3}\cmidrule(lr){3-4}\cmidrule(lr){5-6}\cmidrule(lr){7-8}\cmidrule(lr){9-10}
& Aux. & GloVe & TL & GloVe & TL & GloVe & TL & GloVe & TL \\
\cmidrule(lr){2-3}\cmidrule(lr){3-4}\cmidrule(lr){5-6}\cmidrule(lr){7-8}\cmidrule(lr){9-10}

\multirow{7}{*}{MTL}
& NEG$_{CD}$   &  54.65 \sd{(1.37)} &  62.89 \sd{(1.18)} &  62.50 \sd{(0.42)} &  65.17 \sd{(0.35)} &  65.06 \sd{(2.66)} &  71.04 \sd{(1.13)} &  \textbf{18.88} \sd{(1.17)} &  22.25$^\star$ \sd{(2.00)} \\

& DR  &  53.67 \sd{(0.94)} &  62.29 \sd{(1.32)} &  62.05$^\star$ \sd{(0.32)} &  65.10 \sd{(0.63)} &  \textbf{66.06} \sd{(2.63)} &  71.45 \sd{(1.47)} &  17.03 \sd{(1.12)} &  22.09$^\star$ \sd{(0.70)} \\

&LEX &  \textbf{54.85} \sd{(0.99)} &  62.55 \sd{(1.66)} &  62.14$^\star$ \sd{(0.83)} &  64.65$^\star$ \sd{(0.88)} &  65.89 \sd{(1.32)} &  \textbf{71.77} \sd{(1.88)} &  18.66 \sd{(1.22)} &  22.74 \sd{(1.68)} \\

&NEG$_{SFU}$  &  53.73 \sd{(1.93)} &  62.61 \sd{(1.79)} &  62.34$^\star$ \sd{(0.54)} &  65.00$^\star$ \sd{(0.48)} &  65.82 \sd{(1.31)} &  71.63 \sd{(1.64)} &  17.60 \sd{(0.57)} &  22.30$^\star$ \sd{(1.19)} \\

&SPEC &  51.65 \sd{(2.32)} &  62.03 \sd{(1.14)} &  62.16 \sd{(0.71)} &  64.50$^\star$ \sd{(0.79)} &  65.16 \sd{(1.50)} &  71.51 \sd{(1.16)} &  16.70 \sd{(2.26)} &  22.86$^\star$ \sd{(0.98)} \\

&UPOS &  54.17 \sd{(2.26)} &  62.35 \sd{(0.77)} &  62.79 \sd{(0.37)} &  64.88 \sd{(0.46)} &  65.73 \sd{(1.46)} &  70.38 \sd{(1.63)} &  18.70 \sd{(0.25)} &  23.05$^\star$ \sd{(0.88)} \\

STL   &     &  54.37 \sd{(2.56)} &  \textbf{63.70} \sd{(1.14)} &  \textbf{63.20} \sd{(0.65)} &  \textbf{65.70} \sd{(0.55)} &  65.60 \sd{(1.06)} &  70.68 \sd{(1.53)} &  18.11 \sd{(2.83)} &  \textbf{24.66} \sd{(1.07)} \\
\bottomrule
\end{tabular}
    }
    \caption{The \Fi results for the test split, where the values represent the mean (standard deviation) of five runs with a different random seed. The \textbf{bold} values represent the best performing model for that dataset and embedding. The $^\star$ represent the models that perform statistically significantly worse than the STL model for that dataset and embedding at a 95\% confidence level.}
    \label{tab:Main_Test_Results}
\end{table*}

\textbf{Training:} For the STL and the MTL models, we tune hyperparameters using \textit{AllenTune} \cite{dodge-etal-2019-show} on the Laptop development dataset. We then use the best hyperparameters on the Laptop dataset for all the STL and MTL experiments, in order to reduce hyperparameter search. We follow the result checklist for hyperparameter searches from \cite{dodge-etal-2019-show} (details found in Tables \ref{table:Parameters_STL} and \ref{table:Parameters_MTL} of Appendix \ref{appendix:hyperparameter_search} along with Figure \ref{fig:single_task_tuning_laptop_performance} showing the expected validation scores from the hyperparameter tuning). For the MTL model, a single epoch involves training for one epoch on the auxiliary task and then an epoch on the main task, as previous work has shown training the lower-level task first improves overall results \cite{hashimoto-etal-2017-joint}. In this work, we assume all of the auxiliary training tasks are conceptually lower than TSA.

\textbf{Evaluation:} For all experiments, we run each model five times \cite{reimers-gurevych-2017-reporting} and report the mean and standard derivation. We also take the distribution of the five runs to perform significance testing \cite{reimers2018comparing}, eliminating the need for Bonferroni correction. Following \newcite{dror-etal-2018-hitchhikers}, we use the non-parametric Wilcoxon signed-rank test \cite{wilcoxon1992individual} for the \F metrics and a more powerful parametric Welch’s t-test \cite{welch1947generalization} for the accuracy metric.

\subsection{Results}
We report the \F score for the target extraction (\Fa), macro \F (\Fs) and accuracy score (\accs) for the sentiment classification for all targets that have been correctly identified by the model, and finally the \F score for the full targeted task (\Fi), following \newcite{he-etal-2019-interactive}.
Unlike \newcite{he-etal-2019-interactive}, we do not use any of the samples that contain the \textit{conflict} label on Laptop or Restaurant. The test results for the main \Fi metric are reported in Table \ref{tab:Main_Test_Results}, 
and the other metrics for the test split are reported in Tables \ref{tab:Main_Test_Results_All_Metrics_Laptop_MPQA} and \ref{tab:Main_Test_Results_All_Metrics_Restaurant_MAMS} of Appendix \ref{appendix_main_results}.

The MTL models outperform STL on four of the eight experiments (see Table \ref{tab:Main_Test_Results}), although the STL TL model is significantly better than the majority of MTL models on MPQA. Of the MTL models, NEG$_{CD}$ + GloVe performs best on MPQA (18.88), DR + GloVe is best on Restaurant (66.06), and LEX is the best model on Laptop (54.85) with GloVe and Restaurant (71.77) with TL. The TL models consistently outperform the GloVe models -- by an average of 5.4 percentage points (pp) across all experiments -- and give the best performance on all datasets. 

The results suggest that transfer learning reduces the beneficial effects of MTL. At the same time, the results suggest that MTL does not hurt the STL models, as no STL model is significantly better than all of the MTL models across the datasets and embeddings for the \Fi metric. \footnote{These findings also generalise to the results on the development splits, shown in Tables \ref{tab:Main_Development_Results_All_Metrics_Laptop_MPQA} and \ref{tab:Main_Development_Results_All_Metrics_Restaurant_MAMS} within Appendix \ref{appendix_main_results}.}

\begin{table*}[!h]
    \centering
    \begin{tabular}{cccP{1.2cm}P{1.2cm}P{1.2cm}P{1.2cm}P{1.2cm}P{1.3cm}P{1.2cm}}
\toprule
 &   &  &      NEG$_{CD}$ &      DR &     LEX &     NEG$_{SFU}$ &    SPEC &    UPOS &           STL \\
 
\cmidrule(lr){4-4}\cmidrule(lr){5-5}\cmidrule(lr){6-6}\cmidrule(lr){7-7}\cmidrule(lr){8-8}\cmidrule(lr){9-9}\cmidrule(lr){10-10}

\multirow{8}{*}{\rotatebox[origin=c]{90}{Laptop$_{Neg}$}} & \multirow{2}{*}{sentiment} & GloVe &  \text{\posbox{42.80}} \sd{\sd{(2.48)}} &  \text{\posbox{38.54}}$^\star$ \sd{\sd{(0.98)}} &  \text{\posbox{38.72}}$^\star$ \sd{(3.00)} &  \textbf{\posbox{45.26}} \sd{(1.45)} &  \text{\posbox{41.23}}$^\star$ \sd{(2.90)} &  \text{\posbox{38.92}}$^\star$ \sd{(1.74)} &  38.32$^\star$ \sd{(1.73)} \\
&      & TL &  \textbf{\posbox{48.49}} \sd{(2.32)} &  45.90 \sd{(3.54)} &  45.93 \sd{(2.13)} &  \text{\posbox{47.04}} \sd{(2.93)} &  45.71 \sd{(2.19)} &  46.29 \sd{(2.03)} &  46.50 \sd{(3.30)} \\

\cmidrule(lr){4-4}\cmidrule(lr){5-5}\cmidrule(lr){6-6}\cmidrule(lr){7-7}\cmidrule(lr){8-8}\cmidrule(lr){9-9}\cmidrule(lr){10-10}

& \multirow{2}{*}{extraction} & GloVe &  75.36$^\star$ \sd{(0.91)} &  76.05$^\star$ \sd{(1.20)} &  \textbf{\posbox{78.68}} \sd{(0.97)} &  75.04$^\star$ \sd{(1.92)} &  76.14 \sd{(2.06)} &  \text{\posbox{77.98}} \sd{(1.41)} &  76.52$^\star$ \sd{(1.24)} \\
&      & TL &  \text{\posbox{82.39}} \sd{(1.34)} &  \text{\posbox{82.95}} \sd{(1.36)} &  \textbf{\posbox{83.47}} \sd{(1.26)} &  \text{\posbox{83.25}} \sd{(1.80)} &  \text{\posbox{82.24}}$^\star$ \sd{(1.39)} &  \text{\posbox{82.58}} \sd{(1.58)} &  82.10 \sd{(1.11)} \\

\cmidrule(lr){4-4}\cmidrule(lr){5-5}\cmidrule(lr){6-6}\cmidrule(lr){7-7}\cmidrule(lr){8-8}\cmidrule(lr){9-9}\cmidrule(lr){10-10}

& \multirow{2}{*}{targeted} & GloVe &  \text{\posbox{32.28}} \sd{(2.23)} &  29.30$^\star$ \sd{(0.54)} & \text{\posbox{30.47}}$^\star$   \sd{(2.45)} &  \textbf{\posbox{33.96}} \sd{(1.30)} & \text{\posbox{31.36}$^\star$} \sd{(1.78)} &  \text{\posbox{30.36}}$^\star$ \sd{(1.56)} &  29.33$^\star$ \sd{(1.47)} \\
&      & TL &  \textbf{\posbox{39.95}} \sd{(2.02)} &  38.08 \sd{(3.13)} &  \text{\posbox{38.35}} \sd{(2.01)} &  \text{\posbox{39.18}} \sd{(2.88)} &  37.59 \sd{(1.99)} &  \text{\posbox{38.23}} \sd{(1.89)} &  38.14 \sd{(2.23)} \\ [2ex]

\hline\\ [-1.5ex]

\multirow{8}{*}{\rotatebox[origin=c]{90}{Restaurant$_{Neg}$}} & \multirow{2}{*}{sentiment} & GloVe &  \text{\posbox{53.41}} \sd{(4.28)} &  49.78$^\star$ \sd{(2.10)} &  47.69$^\star$ \sd{(1.19)} &  \textbf{\posbox{56.01}} \sd{(1.07)} &  48.86$^\star$ \sd{(3.94)} &  \text{\posbox{50.58}}$^\star$ \sd{(2.18)} &  49.86$^\star$ \sd{(1.77)} \\
&      & TL &  \text{\posbox{60.69}} \sd{(1.91)} &  \textbf{\posbox{62.61}} \sd{(2.11)} &  \text{\posbox{60.80}} \sd{(3.20)} &  60.45 \sd{(2.04)} &  \text{\posbox{61.70}} \sd{(1.42)} &  60.06 \sd{(2.13)} &  60.66 \sd{(2.24)} \\

\cmidrule(lr){4-4}\cmidrule(lr){5-5}\cmidrule(lr){6-6}\cmidrule(lr){7-7}\cmidrule(lr){8-8}\cmidrule(lr){9-9}\cmidrule(lr){10-10}

& \multirow{2}{*}{extraction} & GloVe &  80.97 \sd{(1.47)} &  \textbf{\posbox{82.22}} \sd{(1.29)} &  \text{\posbox{82.15}} \sd{(0.74)} &  80.74 \sd{(1.58)} &  \text{\posbox{81.53}} \sd{(0.32)} &  \text{\posbox{81.92}} \sd{(0.91)} &  80.97 \sd{(1.14)} \\
&      & TL &  83.04 \sd{(1.26)} &  82.94$^\star$ \sd{(0.97)} &  \textbf{\posbox{84.10}} \sd{(0.86)} &  \text{\posbox{83.94}} \sd{(1.67)} &  83.48 \sd{(1.59)} &  82.33$^\star$ \sd{(1.37)} &  83.50 \sd{(1.16)} \\

\cmidrule(lr){4-4}\cmidrule(lr){5-5}\cmidrule(lr){6-6}\cmidrule(lr){7-7}\cmidrule(lr){8-8}\cmidrule(lr){9-9}\cmidrule(lr){10-10}

& \multirow{2}{*}{targeted} & GloVe &  \text{\posbox{43.28}} \sd{(3.95)} &  \text{\posbox{40.95}}$^\star$ \sd{(2.31)} &  39.19$^\star$ \sd{(1.23)} &  \textbf{\posbox{45.22}} \sd{(0.80)} &  39.85$^\star$ \sd{(3.35)} &  \text{\posbox{41.43}}$^\star$ \sd{(1.87)} &  40.38$^\star$ \sd{(1.82)} \\
&      & TL &  50.40 \sd{(2.03)} &  \textbf{\posbox{51.92}} \sd{(1.64)} &  \text{\posbox{51.15}} \sd{(3.04)} &  \text{\posbox{50.75}} \sd{(2.10)} &  \text{\posbox{51.49}} \sd{(0.86)} &  49.45 \sd{(2.01)} &  50.68 \sd{(2.52)} \\
\bottomrule
\end{tabular}
    \caption{Sentiment (\accs), extraction (\Fa) and full targeted (\Fi) results for Laptop$_{Neg}$ and Restaurant$_{Neg}$ test split, where the values represent the mean (standard deviation) of five runs with a different random seeds. The \textbf{bold} values represent the best model, while \posbox{highlighted} models are those that perform better than the single task baseline. The $^\star$ represents the models that are significantly worse (p $ < 0.05$) than the best performing model on the respective dataset, metric, and embedding.} 
    \label{tab:Negation_Only_Test_Results}
\end{table*}

\begin{table*}[!h]
    \centering
    \begin{tabular}{cccP{1.2cm}P{1.2cm}P{1.2cm}P{1.2cm}P{1.2cm}P{1.3cm}P{1.2cm}}
\toprule
 &   &  &     NEG$_{CD}$ &     DR &     LEX &     NEG$_{SFU}$ &    SPEC &   UPOS &           STL \\

\cmidrule(lr){4-4}\cmidrule(lr){5-5}\cmidrule(lr){6-6}\cmidrule(lr){7-7}\cmidrule(lr){8-8}\cmidrule(lr){9-9}\cmidrule(lr){10-10}

\multirow{8}{*}{\rotatebox[origin=c]{90}{Laptop$_{Spec}$}} & \multirow{2}{*}{sentiment} & GloVe &  34.32 \sd{\sd{(1.86)}} &  \text{\posbox{35.67}} \sd{\sd{(1.00)}} &  \textbf{\posbox{36.75}} \sd{\sd{(1.91)}} &  \text{\posbox{35.98}} \sd{\sd{(2.05)}} &  \text{\posbox{36.74}} \sd{\sd{(1.64)}} &  \text{\posbox{35.57}} \sd{\sd{(1.31)}} &  34.67 \sd{\sd{(1.40)}} \\
&      & TL  &  35.42 \sd{\sd{(3.54)}} &  34.76 \sd{\sd{(1.63)}} &  35.06 \sd{\sd{(1.97)}} &  34.08$^\star$ \sd{\sd{(0.40)}} &  35.03 \sd{\sd{(2.36)}} &  35.01 \sd{\sd{(1.04)}} &  \textbf{35.97} \sd{\sd{(1.45)}} \\

\cmidrule(lr){4-4}\cmidrule(lr){5-5}\cmidrule(lr){6-6}\cmidrule(lr){7-7}\cmidrule(lr){8-8}\cmidrule(lr){9-9}\cmidrule(lr){10-10}

& \multirow{2}{*}{extraction} & GloVe &  74.77$^\star$ \sd{\sd{(1.54)}} &  74.01$^\star$ \sd{\sd{(1.93)}} &  \textbf{\posbox{77.80}} \sd{\sd{(1.34)}} &  \text{\posbox{75.99}} \sd{\sd{(2.48)}} &  73.39$^\star$ \sd{\sd{(1.74)}} &  \text{\posbox{76.80}} \sd{\sd{(0.99)}} &  75.01 \sd{\sd{(1.93)}} \\
&      & TL  &  \text{\posbox{80.11}}$^\star$ \sd{\sd{(1.40)}} &  \text{\posbox{80.77}} \sd{\sd{(1.23)}} &  \text{\posbox{81.47}}$^\star$ \sd{\sd{(0.50)}} &  \textbf{\posbox{83.14}} \sd{\sd{(2.22)}} &  \text{\posbox{81.49}} \sd{\sd{(1.24)}} &  \text{\posbox{81.07}} \sd{\sd{(1.38)}} &  79.84$^\star$ \sd{\sd{(0.58)}} \\

\cmidrule(lr){4-4}\cmidrule(lr){5-5}\cmidrule(lr){6-6}\cmidrule(lr){7-7}\cmidrule(lr){8-8}\cmidrule(lr){9-9}\cmidrule(lr){10-10}

& \multirow{2}{*}{targeted} & GloVe &  25.67$^\star$ \sd{\sd{(1.62)}} &  \text{\posbox{26.39}}$^\star$ \sd{\sd{(0.60)}} &  \textbf{\posbox{28.59}} \sd{\sd{(1.42)}} &  \text{\posbox{27.33}} \sd{\sd{(1.70)}} &  \text{\posbox{26.95}} \sd{\sd{(1.07)}} &  \text{\posbox{27.31}} \sd{\sd{(0.82)}} &  26.01$^\star$ \sd{\sd{(1.26)}} \\
&      & TL  &  28.36 \sd{\sd{(2.81)}} &  28.09 \sd{\sd{(1.68)}} &  28.56 \sd{\sd{(1.60)}} &  28.33 \sd{\sd{(0.52)}} &  28.54 \sd{\sd{(1.83)}} &  28.37 \sd{\sd{(0.77)}} &  \textbf{28.72} \sd{\sd{(1.20)}} \\[2ex]

\hline\\ [-1.5ex]

\multirow{8}{*}{\rotatebox[origin=c]{90}{Restaurant$_{Spec}$}} & \multirow{2}{*}{sentiment} & GloVe &  62.38 \sd{\sd{(3.75)}} &  \text{\posbox{64.01}} \sd{\sd{(2.72)}} &  63.44 \sd{\sd{(2.21)}} &  63.33 \sd{\sd{(1.87)}} &  \textbf{\posbox{64.30}} \sd{\sd{(3.14)}} &  63.15 \sd{\sd{(3.38)}} &  63.94 \sd{\sd{(1.84)}} \\
&      & TL  &  67.23 \sd{\sd{(1.08)}} &  \text{\posbox{68.98}} \sd{\sd{(1.17)}} &  \textbf{\posbox{69.70}} \sd{\sd{(2.51)}} &  67.62 \sd{\sd{(1.58)}} &  66.93 \sd{\sd{(1.79)}} &  68.13 \sd{\sd{(1.25)}} &  68.17 \sd{\sd{(2.44)}} \\

\cmidrule(lr){4-4}\cmidrule(lr){5-5}\cmidrule(lr){6-6}\cmidrule(lr){7-7}\cmidrule(lr){8-8}\cmidrule(lr){9-9}\cmidrule(lr){10-10}

& \multirow{2}{*}{extraction} & GloVe &  75.53 \sd{\sd{(1.03)}} &  \textbf{\posbox{76.40}} \sd{\sd{(1.90)}} &  \text{\posbox{75.75}} \sd{\sd{(1.18)}} &  \text{\posbox{75.66}} \sd{\sd{(1.65)}} &  75.29 \sd{\sd{(0.77)}} &  \text{\posbox{75.87}} \sd{\sd{(0.97)}} &  75.58 \sd{\sd{(1.48)}} \\
&      & TL  &  \text{\posbox{77.92}} \sd{\sd{(1.36)}} &  \text{\posbox{77.84}} \sd{\sd{(0.84)}} &  \textbf{\posbox{79.10}} \sd{\sd{(1.48)}} &  \text{\posbox{78.76}} \sd{\sd{(1.27)}} &  \text{\posbox{78.20}} \sd{\sd{(1.80)}} &  77.15 \sd{\sd{(1.92)}} &  77.61 \sd{\sd{(1.87)}} \\

\cmidrule(lr){4-4}\cmidrule(lr){5-5}\cmidrule(lr){6-6}\cmidrule(lr){7-7}\cmidrule(lr){8-8}\cmidrule(lr){9-9}\cmidrule(lr){10-10}

& \multirow{2}{*}{targeted} & GloVe &  47.14 \sd{\sd{(3.24)}} &  \textbf{\posbox{48.94}} \sd{\sd{(3.06)}} &  48.07 \sd{\sd{(2.22)}} &  47.90 \sd{\sd{(1.25)}} &  \text{\posbox{48.41}} \sd{\sd{(2.48)}} &  47.94 \sd{\sd{(3.14)}} &  48.35 \sd{\sd{(2.32)}} \\
&      & TL  &  52.39 \sd{\sd{(1.18)}} &  \text{\posbox{53.69}} \sd{\sd{(0.69)}} &  \textbf{\posbox{55.15}} \sd{\sd{(2.70)}} &  \text{\posbox{53.25}} \sd{\sd{(1.10)}} &  52.34 \sd{\sd{(1.85)}} &  52.55 \sd{\sd{(1.22)}} &  52.94 \sd{\sd{(2.99)}} \\
\bottomrule
\end{tabular}
    \caption{Sentiment (\accs), extraction (\Fa) and full targeted (\Fi) results for Laptop$_{Spec}$ and Restaurant$_{Spec}$ test split, where the values represent the mean (standard deviation) of five runs with a different random seeds. The \textbf{bold} values represent the best model, while \posbox{highlighted} models are those that perform better than the single task baseline. The $^\star$ represents the models that are significantly worse (p $ < 0.05$) than the best performing model on the respective dataset, metric, and embedding.} 
    \label{tab:Speculation_Only_Test_Results}
\end{table*}

\section{Challenge Dataset Results}

\label{analysis}
In order to isolate the effects of negation and speculation on the results, we test all models trained on the original Laptop and Restaurant datasets on the Laptop$_{Neg}$, Restaurant$_{Neg}$, Laptop$_{Spec}$, and Restaurant$_{Spec}$ test splits. Tables \ref{tab:Negation_Only_Test_Results} and \ref{tab:Speculation_Only_Test_Results} show the results for negation and speculation, respectively. The results for the dev split and the \Fs of the test split are shown in Appendix \ref{appendix_neg_spec_results}.

Firstly, all models perform comparatively worse on the challenge datasets, dropping an average of 24 and 25 pp on \Fi on the negation and speculation data, respectively. Nearly all of this drop comes from poorer classification (\accs, \Fs), while target extraction (\Fa) is relatively stable. This demonstrates the importance of resolving negation and speculation for TSA and the usefulness of the annotated data to determine these effects. 

On Laptop$_{Neg}$ and Restaurant$_{Neg}$ incorporating negation auxiliary tasks gives an average improvement of 3.8 pp on the \Fi metric when using GloVe embeddings. 
More specifically, MTL with negation improves the sentiment classification scores, but does not help extraction. This makes sense conceptually, as negation has little effect on whether or not a word is part of a sentiment target. Instead, jointly learning dependency relations (DR) and full lexical analysis (LEX) improve extraction results. Furthermore, when using TL instead of GloVe embeddings, the best MTL model (NEG$_{SFU}$) does marginally beat the STL TL equivalent on average, indicating that multi-task learning is still able to contribute something to transfer learning.

On Laptop$_{Spec}$ and Restaurant$_{Spec}$ MTL models improve results when using GloVe embeddings, with the additional speculation (SPEC) and dependency relation (DR) data improving the \Fi metric by 0.5 pp and 0.49 pp respectively on average. However, with TL, MTL only leads to benefits on the Restaurant dataset. Unlike the negation data results, the speculation results appear to be helped more by syntactic auxiliary tasks like DR than semantic tasks like NEG$_{CD}$ and to some extent NEG$_{SFU}$.

The best MTL GloVe models on the original datasets (LEX\footnote{The development \Fi result for LEX on the Laptop dataset is worse than STL by 0.05 but for all other \Fi Laptop results LEX is better than STL.} and DR, respectively) also outperform the STL GloVe models on the challenge data, indicating that MTL leads to greater robustness. When comparing the STL model using GloVe and TL on average the model improves by 9.55 pp on the negation dataset compared to 3.65 pp for the speculation suggesting that transfer learning is less effective for speculation.

\section{Conclusion}
In this paper, we have compared the effects of MTL using various auxiliary tasks for TSA and have created a negation and speculation annotated challenge dataset\footnote{\url{https://bit.ly/312kwpP}}
for TSA in order to isolate the effects of MTL. We show that TSA methods are drastically affected by negation and speculation effects in the data. These effects can be similarly reduced by either incorporating auxiliary task information into the model through MTL or through transfer learning. Additionally, MTL of negation can lead to small improvements when combined with transfer learning. Returning to the two original research questions, we can conclude that in general 1) MTL using negation (speculation) as an auxiliary task does make TSA models more robust to negated (speculative) samples and 2) transfer learning seems to incorporate much of the same knowledge. Additionally, incorporating syntactic information as an auxiliary task within MTL creates models that are more robust to both negation and speculation. 

Neither MTL nor TL are currently guarantees for improved performance\footnote{Compare the performance of LEX using GloVe (28.59) to when it uses TL (28.56) in Table \ref{tab:Speculation_Only_Test_Results} for the Laptop dataset.}. Additionally, the results from the challenge datasets indicate that different auxiliary tasks improve the performance of different subtasks of TSA. This may suggest that the target extraction and sentiment classification tasks should not be treated as a collapsed labelling task, as the sentiment and extraction tasks are too dissimilar \cite{hu-etal-2019-open}. Future work should consider using pipeline or joint approaches, where each subtask can be paired with the most beneficial auxiliary tasks. This decoupling could also allow MTL and transfer learning to compliment each other more. 

Finally, in order to improve reproducibility and to encourage further work, we release the code\footnote{\url{https://github.com/jerbarnes/multitask_negation_for_targeted_sentiment}}, dataset, and trained models associated with this paper, hyperparameter search details with compute infrastructure (Appendix \ref{appendix:hyperparameter_search}), number of parameters and runtime details (Appendix \ref{appendix:additional_reproducibility_statistics}), and further detailed dev and test results (appendices \ref{appendix_main_results} and \ref{appendix_neg_spec_results}), in line with the result checklist from \newcite{dodge-etal-2019-show}.

\label{sec:summary}

\section*{Acknowledgements}
This work has been carried out as part of the SANT project (Sentiment Analysis for Norwegian Text), funded by the Research Council of Norway (grant number 270908). Andrew has been funded by Lancaster University by an EPSRC Doctoral Training Grant. The authors thank the UCREL research centre for hosting the models created from this research.

\bibliography{lit}
\bibliographystyle{acl_natbib}

\clearpage
\appendix


\section{Class Distribution of the Sentiment Datasets}
\label{appendix_label_distribution}

\begin{table*}[h!]
\centering
\resizebox{\textwidth}{!}{%
\begin{tabular}{lcccccccccccc}
\toprule
{} & \multicolumn{4}{c}{\textbf{Train}} & \multicolumn{4}{c}{\textbf{Dev}} & \multicolumn{4}{c}{\textbf{Test}} \\
\cmidrule(lr){2-5}\cmidrule(lr){6-9}\cmidrule(lr){10-13}
{} &    pos &    neu &    neg &  both &    pos &    neu &    neg & both &    pos &    neu &    neg & both \\
\cmidrule(lr){2-2}\cmidrule(lr){3-3}\cmidrule(lr){4-4}\cmidrule(lr){5-5}\cmidrule(lr){6-6}\cmidrule(lr){7-7}\cmidrule(lr){8-8}\cmidrule(lr){9-9}\cmidrule(lr){10-10}\cmidrule(lr){11-11}\cmidrule(lr){12-12}\cmidrule(lr){13-13}
Laptop          &   19.9 &  43.2 &  36.9 &  - & 18.0 &  40.6 &  41.4 &  - &     26.0 &  53.5 &  20.5 &  - \\
Laptop$_{Neg}$      &  - &   - &   - &  - &  17.1 &  35.4 &  47.5 &  - &    26.7 &  23.3 &  50.0 &  - \\
Laptop$_{Spec}$     &  - &   - &   - &  - &  50.7 &  16.2 &  33.1 &  - &    38.2 &  20.5 &  41.4 &  - \\
Restaurant      &   15.8 &  60.0 &  24.2 &  - & 12.3 &  65.2 &  22.5 &  - &   11.5 &  66.6 &  21.9 &  - \\
Restaurant$_{Neg}$  &      - &   - &   - &  - &  16.4 &  32.5 &  51.1 &  - &  15.0 &  32.2 &  52.8 &  - \\
Restaurant$_{Spec}$ &  - &   - &   - &  - &  30.0 &  29.0 &  41.0 &  - &    16.9 &  39.7 &  43.5 &  - \\
MAMS            &   45.1 &  30.2 &  24.7 &  - &  45.5 &  30.3 &  24.3 &  - &    45.5 &  29.9 &  24.6 &  - \\
MPQA            &   13.3 &  43.9 &  39.1 &  3.7 &  17.0 &  42.5 &  37.0 &  3.5 &   19.2 &  33.2 &  41.4 &  6.3 \\
\bottomrule
\end{tabular}}
\caption{Sentiment class distribution statistics as a percentage of the number of targets (samples), for the sentiment datasets used in the experiments. pos, neu, neg, and both represent the sentiment classes positive, neutral, negative, and both respectively.}
\label{table:appendix_label_distribution_stats}
\end{table*}

\clearpage

\section{Examples of Auxiliary Tasks}
\label{appendix_auxiliary_tasks}

\begin{table*}[!h]
\centering
\begin{tabular}{llllllll}
\toprule
& you & might & not & like & the & service \\
NEG$_{CD}$ & B$_{scope}$ & I$_{scope}$ & B$_{cue}$ & B$_{scope}$ & I$_{scope}$ & I$_{scope}$   \\
NEG$_{SFU}$ & B$_{scope}$ & I$_{scope}$ & B$_{cue}$ & B$_{scope}$ & I$_{scope}$ & I$_{scope}$  \\
SPEC & B$_{scope}$ & B$_{cue}$ & B$_{scope}$ & I$_{scope}$ & I$_{scope}$ & I$_{scope}$\\
UPOS & PRON & AUX & PART & VERB & DET & NOUN\\
DR & nsubj & aux & advmod & root & det & obj\\
LEX & O$_{PRON}$ & O$_{AUX}$ & O$_{ADV}$ & B$_{V-v.emotion}$ & O$_{DET}$ & B$_{N-n.ACT}$ \\
\bottomrule
\end{tabular}
\caption{A toy example sentence with the labels from each auxiliary task}
\end{table*}

\clearpage

\section{Additional Main Result Tables}
\label{appendix_main_results}

\begin{table*}[!h]
    \centering
    \begin{tabular}{cccP{1.2cm}P{1.2cm}P{1.2cm}P{1.2cm}P{1.2cm}P{1.3cm}P{1.2cm}}
\toprule
 &   &  &      NEG$_{CD}$ &      DR &     LEX &     NEG$_{SFU}$ &    SPEC &    UPOS &           STL \\

\cmidrule(lr){4-4}\cmidrule(lr){5-5}\cmidrule(lr){6-6}\cmidrule(lr){7-7}\cmidrule(lr){8-8}\cmidrule(lr){9-9}\cmidrule(lr){10-10}

\multirow{8}{*}{\rotatebox[origin=c]{90}{Laptop}} & \multirow{2}{*}{acc-s} & GloVe &  \textbf{\posbox{71.90}} \sd{\sd{(1.32)}} &  70.66 \sd{\sd{(1.55)}} &  70.36 \sd{\sd{(2.24)}} &  70.30 \sd{\sd{(1.86)}} &  68.11$^\star$ \sd{\sd{(2.19)}} &  69.60$^\star$ \sd{\sd{(1.98)}} &  70.80 \sd{\sd{(2.02)}} \\
     &      & TL  &  75.30 \sd{\sd{(0.54)}} &  74.75 \sd{\sd{(1.14)}} &  74.56 \sd{\sd{(1.49)}} &  74.36 \sd{\sd{(1.74)}} &  74.47$^\star$ \sd{\sd{(0.82)}} &  74.70$^\star$ \sd{\sd{(1.02)}} &  \textbf{76.85} \sd{\sd{(1.96)}} \\

\cmidrule(lr){4-4}\cmidrule(lr){5-5}\cmidrule(lr){6-6}\cmidrule(lr){7-7}\cmidrule(lr){8-8}\cmidrule(lr){9-9}\cmidrule(lr){10-10}

& \multirow{2}{*}{F1-s} & GloVe &  \textbf{\posbox{65.00}} \sd{\sd{(1.36)}} &  \posbox{63.19} \sd{\sd{(2.32)}} &  \posbox{63.07} \sd{\sd{(3.51)}} &  \posbox{62.60}$^\star$ \sd{\sd{(2.74)}} &  59.83$^\star$ \sd{\sd{(2.46)}} &  61.51 \sd{\sd{(2.66)}} &  61.90 \sd{\sd{(3.32)}} \\
&      & TL  &  66.92 \sd{\sd{(2.41)}} &  67.76 \sd{\sd{(1.75)}} &  67.26 \sd{\sd{(2.27)}} &  66.00 \sd{\sd{(3.03)}} &  66.92$^\star$ \sd{\sd{(1.21)}} &  66.63$^\star$ \sd{\sd{(1.61)}} &  \textbf{69.91} \sd{\sd{(2.72)}} \\

\cmidrule(lr){4-4}\cmidrule(lr){5-5}\cmidrule(lr){6-6}\cmidrule(lr){7-7}\cmidrule(lr){8-8}\cmidrule(lr){9-9}\cmidrule(lr){10-10}

& \multirow{2}{*}{extraction} & GloVe &  76.00$^\star$ \sd{\sd{(0.99)}} &  75.98$^\star$ \sd{\sd{(1.17)}} &  \textbf{\posbox{77.99}} \sd{\sd{(1.14)}} &  76.43 \sd{\sd{(1.57)}} &  75.81 \sd{\sd{(1.57)}} &  \posbox{77.81} \sd{\sd{(1.10)}} &  76.76 \sd{\sd{(1.69)}} \\
&      & TL  &  \posbox{83.51} \sd{\sd{(1.09)}} &  \posbox{83.32} \sd{\sd{(0.94)}} &  \posbox{83.88} \sd{\sd{(0.88)}} &  \textbf{\posbox{84.21}} \sd{\sd{(1.81)}} &  \posbox{83.29} \sd{\sd{(1.15)}} &  \posbox{83.48} \sd{\sd{(1.30)}} &  82.90 \sd{\sd{(0.72)}} \\

\cmidrule(lr){4-4}\cmidrule(lr){5-5}\cmidrule(lr){6-6}\cmidrule(lr){7-7}\cmidrule(lr){8-8}\cmidrule(lr){9-9}\cmidrule(lr){10-10}

& \multirow{2}{*}{targeted} & GloVe &  \posbox{54.65} \sd{\sd{(1.37)}} &  53.67 \sd{\sd{(0.94)}} &  \textbf{\posbox{54.85}} \sd{\sd{(0.99)}} &  53.73 \sd{\sd{(1.93)}} &  51.65$^\star$ \sd{\sd{(2.32)}} &  54.17 \sd{\sd{(2.26)}} &  54.37 \sd{\sd{(2.56)}} \\
&      & TL  &  62.89 \sd{\sd{(1.18)}} &  62.29 \sd{\sd{(1.32)}} &  62.55 \sd{\sd{(1.66)}} &  62.61 \sd{\sd{(1.79)}} &  62.03 \sd{\sd{(1.14)}} &  62.35 \sd{\sd{(0.77)}} &  \textbf{63.70} \sd{\sd{(1.14)}} \\[2ex]

\hline\\ [-1.5ex]

\multirow{8}{*}{\rotatebox[origin=c]{90}{MPQA}} & \multirow{2}{*}{acc-s} & GloVe &  \posbox{\textbf{78.18}} \sd{\sd{(2.72)}} &  \posbox{74.37} \sd{\sd{(3.47)}} &  \posbox{75.94} \sd{\sd{(3.02)}} &  \posbox{77.38} \sd{\sd{(4.91)}} &  72.82$^\star$ \sd{\sd{(3.88)}} &  \posbox{73.83}$^\star$ \sd{\sd{(2.30)}} &  73.01$^\star$ \sd{\sd{(3.41)}} \\
&      & TL  &  \posbox{71.84} \sd{\sd{(3.46)}} &  \posbox{72.01} \sd{\sd{(3.01)}} &  \posbox{\textbf{73.08}} \sd{\sd{(4.01)}} &  \posbox{70.96} \sd{\sd{(2.03)}} &  \posbox{72.79} \sd{\sd{(3.13)}} &  \posbox{72.61} \sd{\sd{(3.84)}} &  70.47 \sd{\sd{(1.51)}} \\

\cmidrule(lr){4-4}\cmidrule(lr){5-5}\cmidrule(lr){6-6}\cmidrule(lr){7-7}\cmidrule(lr){8-8}\cmidrule(lr){9-9}\cmidrule(lr){10-10}

& \multirow{2}{*}{F1-s} & GloVe &  \textbf{\posbox{42.03}} \sd{\sd{(1.50)}} &  \posbox{39.96} \sd{\sd{(1.66)}} &  \posbox{40.58} \sd{\sd{(1.28)}} &  \posbox{41.16} \sd{\sd{(2.32)}} &  \posbox{39.19} \sd{\sd{(1.94)}} &  \posbox{39.90}$^\star$ \sd{\sd{(1.06)}} &  39.00 \sd{\sd{(1.86)}} \\
&      & TL  &  \posbox{39.92} \sd{\sd{(1.15)}} &  \posbox{40.27} \sd{\sd{(0.80)}} &  \textbf{\posbox{41.13}} \sd{\sd{(3.05)}} &  39.17 \sd{\sd{(0.79)}} &  \posbox{39.84} \sd{\sd{(1.56)}} &  \posbox{39.90} \sd{\sd{(1.66)}} &  39.25 \sd{\sd{(0.68)}} \\

\cmidrule(lr){4-4}\cmidrule(lr){5-5}\cmidrule(lr){6-6}\cmidrule(lr){7-7}\cmidrule(lr){8-8}\cmidrule(lr){9-9}\cmidrule(lr){10-10}

& \multirow{2}{*}{extraction} & GloVe &  24.17 \sd{\sd{(1.44)}} &  22.93$^\star$ \sd{\sd{(1.57)}} &  24.58 \sd{\sd{(1.36)}} &  22.84$^\star$ \sd{\sd{(1.58)}} &  22.90$^\star$ \sd{\sd{(2.49)}} &  \posbox{\textbf{25.34}} \sd{\sd{(0.46)}} &  24.77 \sd{\sd{(3.55)}} \\
&      & TL  &  30.98$^\star$ \sd{\sd{(2.40)}} &  30.71$^\star$ \sd{\sd{(1.10)}} &  31.19$^\star$ \sd{\sd{(2.69)}} &  31.41$^\star$ \sd{\sd{(1.16)}} &  31.41$^\star$ \sd{\sd{(0.51)}} &  31.88$^\star$ \sd{\sd{(2.62)}} &  \textbf{34.99} \sd{\sd{(1.10)}} \\

\cmidrule(lr){4-4}\cmidrule(lr){5-5}\cmidrule(lr){6-6}\cmidrule(lr){7-7}\cmidrule(lr){8-8}\cmidrule(lr){9-9}\cmidrule(lr){10-10}

& \multirow{2}{*}{targeted} & GloVe &  \posbox{\textbf{18.88}} \sd{\sd{(1.17)}} &  17.03$^\star$ \sd{\sd{(1.12)}} &  \posbox{18.66} \sd{\sd{(1.22)}} &  17.60$^\star$ \sd{\sd{(0.57)}} &  16.70$^\star$ \sd{\sd{(2.26)}} &  \posbox{18.70} \sd{\sd{(0.25)}} &  18.11 \sd{\sd{(2.83)}} \\
&      & TL  &  22.25$^\star$ \sd{\sd{(2.00)}} &  22.09$^\star$ \sd{\sd{(0.70)}} &  22.74 \sd{\sd{(1.68)}} &  22.30$^\star$ \sd{\sd{(1.19)}} &  22.86$^\star$ \sd{\sd{(0.98)}} &  23.05$^\star$ \sd{\sd{(0.88)}} &  \textbf{24.66} \sd{\sd{(1.07)}} \\

\bottomrule
\end{tabular}
    \caption{\accs, \Fs, extraction (\Fa) and full targeted (\Fi) results for Laptop and MPQA \textbf{test split}, where the values represent the mean (standard deviation) of five runs with a different random seed. The \textbf{bold} values represent the best model, while \posbox{highlighted} models are those that perform better than the single task baseline. The $^\star$ represent the models that are statistically significantly worse than the best performing model on the respective dataset, metric and TL at a 95\% confidence level.}
    \label{tab:Main_Test_Results_All_Metrics_Laptop_MPQA}
\end{table*}

\begin{table*}[!h]
    \centering
    \begin{tabular}{cccP{1.2cm}P{1.2cm}P{1.2cm}P{1.2cm}P{1.2cm}P{1.3cm}P{1.2cm}}
\toprule
 &   &  &      NEG$_{CD}$ &      DR &     LEX &     NEG$_{SFU}$ &    SPEC &    UPOS &           STL \\

\cmidrule(lr){4-4}\cmidrule(lr){5-5}\cmidrule(lr){6-6}\cmidrule(lr){7-7}\cmidrule(lr){8-8}\cmidrule(lr){9-9}\cmidrule(lr){10-10}

\multirow{8}{*}{\rotatebox[origin=c]{90}{MAMS}} & \multirow{2}{*}{acc-s} & GloVe &  \posbox{\textbf{81.72}} \sd{\sd{(1.12)}} &  81.42 \sd{\sd{(0.32)}} &  81.24 \sd{\sd{(0.44)}} &  81.10 \sd{\sd{(0.46)}} &  80.90 \sd{\sd{(0.70)}} &  81.58 \sd{\sd{(0.96)}} &  81.70 \sd{\sd{(0.79)}} \\
&      & TL  &  84.59 \sd{\sd{(0.50)}} &  \posbox{\textbf{84.81}} \sd{\sd{(0.78)}} &  83.99$^\star$ \sd{\sd{(0.84)}} &  83.73 \sd{\sd{(0.68)}} &  84.28 \sd{\sd{(0.39)}} &  83.90$^\star$ \sd{\sd{(0.51)}} &  84.67 \sd{\sd{(0.56)}} \\

\cmidrule(lr){4-4}\cmidrule(lr){5-5}\cmidrule(lr){6-6}\cmidrule(lr){7-7}\cmidrule(lr){8-8}\cmidrule(lr){9-9}\cmidrule(lr){10-10}

& \multirow{2}{*}{F1-s} & GloVe &  \posbox{\textbf{81.05}} \sd{\sd{(1.15)}} &  80.71 \sd{\sd{(0.43)}} &  80.53 \sd{\sd{(0.60)}} &  80.39 \sd{\sd{(0.69)}} &  80.22 \sd{\sd{(0.80)}} &  80.81 \sd{\sd{(1.01)}} &  80.94 \sd{\sd{(0.88)}} \\
&      & TL  &  84.24 \sd{\sd{(0.51)}} &  \posbox{\textbf{84.39}} \sd{\sd{(0.80)}} &  83.59$^\star$ \sd{\sd{(0.87)}} &  83.30 \sd{\sd{(0.67)}} &  83.89 \sd{\sd{(0.32)}} &  83.44$^\star$ \sd{\sd{(0.60)}} &  84.24 \sd{\sd{(0.58)}} \\

\cmidrule(lr){4-4}\cmidrule(lr){5-5}\cmidrule(lr){6-6}\cmidrule(lr){7-7}\cmidrule(lr){8-8}\cmidrule(lr){9-9}\cmidrule(lr){10-10}

& \multirow{2}{*}{extraction} & GloVe &  76.49 \sd{\sd{(0.99)}} &  76.21$^\star$ \sd{\sd{(0.39)}} &  76.49 \sd{\sd{(1.06)}} &  76.87 \sd{\sd{(0.64)}} &  76.83 \sd{\sd{(0.48)}} &  76.97 \sd{\sd{(0.54)}} &  \textbf{77.36} \sd{\sd{(0.19)}} \\
&      & TL  &  77.04 \sd{\sd{(0.35)}} &  76.76$^\star$ \sd{\sd{(0.13)}} &  76.98 \sd{\sd{(0.84)}} &  \posbox{\textbf{77.64}} \sd{\sd{(0.36)}} &  76.54$^\star$ \sd{\sd{(0.79)}} &  77.33 \sd{\sd{(0.61)}} &  77.59 \sd{\sd{(0.35)}} \\

\cmidrule(lr){4-4}\cmidrule(lr){5-5}\cmidrule(lr){6-6}\cmidrule(lr){7-7}\cmidrule(lr){8-8}\cmidrule(lr){9-9}\cmidrule(lr){10-10}

& \multirow{2}{*}{targeted} & GloVe &  62.50 \sd{\sd{(0.42)}} &  62.05$^\star$ \sd{\sd{(0.32)}} &  62.14$^\star$ \sd{\sd{(0.83)}} &  62.34$^\star$ \sd{\sd{(0.54)}} &  62.16 \sd{\sd{(0.71)}} &  62.79 \sd{\sd{(0.37)}} &  \textbf{63.20} \sd{\sd{(0.65)}} \\
&      & TL  &  65.17 \sd{\sd{(0.35)}} &  65.10 \sd{\sd{(0.63)}} &  64.65$^\star$ \sd{\sd{(0.88)}} &  65.00$^\star$ \sd{\sd{(0.48)}} &  64.50$^\star$ \sd{\sd{(0.79)}} &  64.88 \sd{\sd{(0.46)}} &  \textbf{65.70} \sd{\sd{(0.55)}} \\[2ex]

\hline\\ [-1.5ex]

\multirow{8}{*}{\rotatebox[origin=c]{90}{Restaurant}} & \multirow{2}{*}{acc-s} & GloVe &  83.02 \sd{\sd{(1.82)}} &  83.23 \sd{\sd{(1.69)}} &  83.26 \sd{\sd{(0.89)}} &  \posbox{\textbf{83.80}} \sd{\sd{(0.78)}} &  83.01 \sd{\sd{(1.16)}} &  83.36 \sd{\sd{(1.09)}} &  83.65 \sd{\sd{(0.48)}} \\
&      & TL  &  87.40 \sd{\sd{(0.67)}} &  \posbox{\textbf{87.63}} \sd{\sd{(0.76)}} &  \posbox{87.37} \sd{\sd{(0.90)}} &  87.26 \sd{\sd{(0.96)}} &  \posbox{87.36} \sd{\sd{(0.48)}} &  87.00$^\star$ \sd{\sd{(0.56)}} &  87.32 \sd{\sd{(0.66)}} \\

\cmidrule(lr){4-4}\cmidrule(lr){5-5}\cmidrule(lr){6-6}\cmidrule(lr){7-7}\cmidrule(lr){8-8}\cmidrule(lr){9-9}\cmidrule(lr){10-10}

& \multirow{2}{*}{F1-s} & GloVe &  66.75 \sd{\sd{(3.75)}} &  67.79 \sd{\sd{(3.00)}} &  67.59 \sd{\sd{(1.39)}} &  67.75 \sd{\sd{(1.92)}} &  67.35 \sd{\sd{(3.02)}} &  67.13 \sd{\sd{(2.31)}} &  \textbf{68.00} \sd{\sd{(1.61)}} \\
&      & TL  &  72.27 \sd{\sd{(1.14)}} &  72.96 \sd{\sd{(1.79)}} &  \posbox{73.73} \sd{\sd{(2.60)}} &  72.12 \sd{\sd{(2.30)}} &  \posbox{\textbf{73.90}} \sd{\sd{(2.82)}} &  71.61 \sd{\sd{(1.13)}} &  73.47 \sd{\sd{(1.10)}}\\

\cmidrule(lr){4-4}\cmidrule(lr){5-5}\cmidrule(lr){6-6}\cmidrule(lr){7-7}\cmidrule(lr){8-8}\cmidrule(lr){9-9}\cmidrule(lr){10-10}

& \multirow{2}{*}{extraction} & GloVe &  78.33 \sd{\sd{(1.55)}} &  \posbox{\textbf{79.34}} \sd{\sd{(1.60)}} &  \posbox{79.13} \sd{\sd{(0.93)}} &  \posbox{78.53} \sd{\sd{(0.96)}} &  \posbox{78.48} \sd{\sd{(0.81)}} &  \posbox{78.84} \sd{\sd{(0.78)}} &  78.42 \sd{\sd{(0.85)}} \\
&      & TL  &  \posbox{81.27} \sd{\sd{(0.90)}} &  \posbox{81.53} \sd{\sd{(1.01)}} &  \posbox{\textbf{82.13}} \sd{\sd{(1.35)}} &  \posbox{82.08} \sd{\sd{(1.09)}} &  \posbox{81.85} \sd{\sd{(1.22)}} &  80.89$^\star$ \sd{\sd{(1.37)}} &  80.94 \sd{\sd{(1.18)}} \\

\cmidrule(lr){4-4}\cmidrule(lr){5-5}\cmidrule(lr){6-6}\cmidrule(lr){7-7}\cmidrule(lr){8-8}\cmidrule(lr){9-9}\cmidrule(lr){10-10}

& \multirow{2}{*}{targeted} & GloVe &  65.06 \sd{\sd{(2.66)}} &  \posbox{\textbf{66.06}} \sd{\sd{(2.63)}} &  \posbox{65.89} \sd{\sd{(1.32)}} &  \posbox{65.82} \sd{\sd{(1.31)}} &  65.16 \sd{\sd{(1.50)}} &  \posbox{65.73} \sd{\sd{(1.46)}} &  65.60 \sd{\sd{(1.06)}} \\
&      & TL  &  \posbox{71.04} \sd{\sd{(1.13)}} &  \posbox{71.45} \sd{\sd{(1.47)}} &  \posbox{\textbf{71.77}} \sd{\sd{(1.88)}} &  \posbox{71.63} \sd{\sd{(1.64)}} &  \posbox{71.51} \sd{\sd{(1.16)}} &  70.38 \sd{\sd{(1.63)}} &  70.68 \sd{\sd{(1.53)}} \\

\bottomrule
\end{tabular}
    \caption{\accs, \Fs, extraction (\Fa) and full targeted (\Fi) results for MAMS and Restaurant \textbf{test split}, where the values represent the mean (standard deviation) of five runs with a different random seed. The \textbf{bold} values represent the best model, while \posbox{highlighted} models are those that perform better than the single task baseline. The $^\star$ represent the models that are statistically significantly worse than the best performing model on the respective dataset, metric and TL at a 95\% confidence level.}
    \label{tab:Main_Test_Results_All_Metrics_Restaurant_MAMS}
\end{table*}

\begin{table*}[!h]
    \centering
    \begin{tabular}{cccP{1.2cm}P{1.2cm}P{1.2cm}P{1.2cm}P{1.2cm}P{1.3cm}P{1.2cm}}
\toprule
 &   &  &      NEG$_{CD}$ &      DR &     LEX &     NEG$_{SFU}$ &    SPEC &    UPOS &           STL \\

\cmidrule(lr){4-4}\cmidrule(lr){5-5}\cmidrule(lr){6-6}\cmidrule(lr){7-7}\cmidrule(lr){8-8}\cmidrule(lr){9-9}\cmidrule(lr){10-10}

\multirow{8}{*}{\rotatebox[origin=c]{90}{Laptop}} & \multirow{2}{*}{acc-s} & GloVe &  \posbox{77.47} \sd{\sd{(1.43)}} &   \posbox{77.53} \sd{\sd{(0.69)}} &  76.52 \sd{\sd{(1.38)}} &   \posbox{\textbf{78.34}} \sd{\sd{(2.23)}} &  \posbox{77.32} \sd{\sd{(1.06)}} &  76.67 \sd{\sd{(1.77)}} &  77.22 \sd{\sd{(1.05)}} \\
&      & TL  &  81.22 \sd{\sd{(1.42)}} &   79.63$^\star$ \sd{\sd{(1.56)}} &  80.54 \sd{\sd{(2.27)}} &   80.05$^\star$ \sd{\sd{(1.28)}} &  80.87 \sd{\sd{(1.64)}} &  79.31$^\star$ \sd{\sd{(1.12)}} &  \textbf{81.89} \sd{\sd{(1.07)}} \\

\cmidrule(lr){4-4}\cmidrule(lr){5-5}\cmidrule(lr){6-6}\cmidrule(lr){7-7}\cmidrule(lr){8-8}\cmidrule(lr){9-9}\cmidrule(lr){10-10}

& \multirow{2}{*}{F1-s} & GloVe &  \posbox{71.55} \sd{\sd{(0.81)}} &   \posbox{70.75} \sd{\sd{(2.75)}} &  \posbox{70.61} \sd{\sd{(2.46)}} &   \posbox{\textbf{72.04}} \sd{\sd{(3.85)}} &  \posbox{70.70} \sd{\sd{(1.83)}} &  \posbox{67.82} \sd{\sd{(1.20)}} &  67.27 \sd{\sd{(2.80)}} \\
&      & TL  &  74.53 \sd{\sd{(3.09)}} &   74.84 \sd{\sd{(1.87)}} &  74.89 \sd{\sd{(3.32)}} &   73.70 \sd{\sd{(1.98)}} &  75.22 \sd{\sd{(2.03)}} &  71.90$^\star$ \sd{\sd{(1.77)}} &  \textbf{75.33} \sd{\sd{(1.61)}} \\

\cmidrule(lr){4-4}\cmidrule(lr){5-5}\cmidrule(lr){6-6}\cmidrule(lr){7-7}\cmidrule(lr){8-8}\cmidrule(lr){9-9}\cmidrule(lr){10-10}

& \multirow{2}{*}{extraction} & GloVe &  74.87 \sd{\sd{(1.15)}} &   73.55 \sd{\sd{(2.09)}} &  \posbox{\textbf{75.70}} \sd{\sd{(0.96)}} &   74.81 \sd{\sd{(1.18)}} &  74.75 \sd{\sd{(0.57)}} &  74.15$^\star$ \sd{\sd{(1.84)}} &  75.06 \sd{\sd{(1.06)}} \\
&      & TL  &  80.82 \sd{\sd{(1.35)}} &   80.89 \sd{\sd{(0.83)}} &  80.39 \sd{\sd{(1.02)}} &   \textbf{\posbox{81.86}} \sd{\sd{(1.24)}} &  80.05$^\star$ \sd{\sd{(0.59)}} &  \posbox{81.37} \sd{\sd{(0.99)}} &  81.23 \sd{\sd{(0.82)}} \\

\cmidrule(lr){4-4}\cmidrule(lr){5-5}\cmidrule(lr){6-6}\cmidrule(lr){7-7}\cmidrule(lr){8-8}\cmidrule(lr){9-9}\cmidrule(lr){10-10}

& \multirow{2}{*}{targeted} & GloVe &  \posbox{57.99} \sd{\sd{(0.69)}} &   57.02 \sd{\sd{(1.53)}} &  57.92 \sd{\sd{(1.08)}} &   \posbox{\textbf{58.62}} \sd{\sd{(2.19)}} &  57.80 \sd{\sd{(1.10)}} &  56.82 \sd{\sd{(0.71)}} &  57.97 \sd{\sd{(1.24)}} \\
&      & TL  &  65.62$^\star$ \sd{\sd{(0.76)}} &   64.42 \sd{\sd{(1.64)}} &  64.73 \sd{\sd{(1.30)}} &   65.52$^\star$ \sd{\sd{(0.52)}} &  64.72$^\star$ \sd{\sd{(0.97)}} &  64.55$^\star$ \sd{\sd{(1.53)}} &  \textbf{66.51} \sd{\sd{(0.43)}} \\[2ex]

\hline\\ [-1.5ex]

\multirow{8}{*}{\rotatebox[origin=c]{90}{MPQA}} & \multirow{2}{*}{acc-s} & GloVe &  \posbox{87.75} \sd{\sd{(3.15)}} &   \posbox{88.65} \sd{\sd{(4.20)}} &  \posbox{87.64} \sd{\sd{(3.71)}} &   \posbox{\textbf{89.11}} \sd{\sd{(3.29)}} &  \posbox{86.85} \sd{\sd{(1.46)}} &  \posbox{88.16} \sd{\sd{(3.20)}} &  85.29$^\star$ \sd{\sd{(0.89)}} \\
&      & TL  &  \posbox{88.63} \sd{\sd{(1.83)}} &   \posbox{\textbf{90.08}} \sd{\sd{(2.94)}} &  87.23 \sd{\sd{(1.80)}} &   85.62 \sd{\sd{(4.14)}} &  \posbox{88.71} \sd{\sd{(2.89)}} &  \posbox{88.75} \sd{\sd{(1.56)}} &  88.01 \sd{\sd{(1.36)}} \\

\cmidrule(lr){4-4}\cmidrule(lr){5-5}\cmidrule(lr){6-6}\cmidrule(lr){7-7}\cmidrule(lr){8-8}\cmidrule(lr){9-9}\cmidrule(lr){10-10}

& \multirow{2}{*}{F1-s} & GloVe &  \posbox{54.18} \sd{\sd{(8.15)}} &  \posbox{\textbf{59.87}} \sd{\sd{(14.46)}} &  \posbox{51.51} \sd{\sd{(8.83)}} &  \posbox{56.39} \sd{\sd{(10.28)}} &  \posbox{48.09} \sd{\sd{(4.83)}} &  \posbox{52.32} \sd{\sd{(9.96)}} &  45.92 \sd{\sd{(3.06)}} \\
&      & TL  &  52.83$^\star$ \sd{\sd{(3.40)}} &   \posbox{55.80} \sd{\sd{(5.28)}} &  \posbox{\textbf{59.03}} \sd{\sd{(5.99)}} &   54.48 \sd{\sd{(9.33)}} &  \posbox{56.55} \sd{\sd{(3.28)}} &  53.82 \sd{\sd{(7.01)}} &  55.74 \sd{\sd{(6.52)}} \\

\cmidrule(lr){4-4}\cmidrule(lr){5-5}\cmidrule(lr){6-6}\cmidrule(lr){7-7}\cmidrule(lr){8-8}\cmidrule(lr){9-9}\cmidrule(lr){10-10}

& \multirow{2}{*}{extraction} & GloVe &  \posbox{20.68} \sd{\sd{(0.65)}} &   \posbox{21.00} \sd{\sd{(1.73)}} &  \posbox{20.78} \sd{\sd{(1.52)}} &   \posbox{20.48}$^\star$ \sd{\sd{(0.85)}} &  19.54$^\star$ \sd{\sd{(2.18)}} &  \posbox{\textbf{21.73}} \sd{\sd{(0.74)}} &  20.11 \sd{\sd{(2.38)}} \\
&      & TL  &  32.33 \sd{\sd{(3.71)}} &   30.39$^\star$ \sd{\sd{(1.09)}} &  31.75 \sd{\sd{(1.78)}} &   32.18 \sd{\sd{(1.29)}} &  30.65$^\star$ \sd{\sd{(1.52)}} &  31.00$^\star$ \sd{\sd{(1.92)}} &  \textbf{33.38} \sd{\sd{(0.67)}} \\

\cmidrule(lr){4-4}\cmidrule(lr){5-5}\cmidrule(lr){6-6}\cmidrule(lr){7-7}\cmidrule(lr){8-8}\cmidrule(lr){9-9}\cmidrule(lr){10-10}

& \multirow{2}{*}{targeted} & GloVe &  \posbox{18.14} \sd{\sd{(0.58)}} &   \posbox{18.57} \sd{\sd{(1.13)}} &  \posbox{18.18} \sd{\sd{(1.10)}} &   \posbox{18.23} \sd{\sd{(0.55)}} &  16.94$^\star$ \sd{\sd{(1.66)}} &  \posbox{\textbf{19.16}} \sd{\sd{(0.89)}} &  17.16 \sd{\sd{(2.07)}} \\
&      & TL  &  28.60 \sd{\sd{(2.82)}} &   27.35$^\star$ \sd{\sd{(0.72)}} &  27.67$^\star$ \sd{\sd{(1.29)}} &   27.53$^\star$ \sd{\sd{(1.30)}} &  27.15$^\star$ \sd{\sd{(0.92)}} &  27.49$^\star$ \sd{\sd{(1.38)}} &  \textbf{29.39} \sd{\sd{(0.95)}} \\

\bottomrule
\end{tabular}
    \caption{\accs, \Fs, extraction (\Fa) and full targeted (\Fi) results for Laptop and MPQA \textbf{development split}, where the values represent the mean (standard deviation) of five runs with a different random seed. The \textbf{bold} values represent the best model, while \posbox{highlighted} models are those that perform better than the single task baseline. The $^\star$ represent the models that are statistically significantly worse than the best performing model on the respective dataset, metric and TL at a 95\% confidence level.}
    \label{tab:Main_Development_Results_All_Metrics_Laptop_MPQA}
\end{table*}

\begin{table*}[!h]
    \centering
    \begin{tabular}{cccP{1.2cm}P{1.2cm}P{1.2cm}P{1.2cm}P{1.2cm}P{1.3cm}P{1.2cm}}
\toprule
 &   &  &      NEG$_{CD}$ &      DR &     LEX &     NEG$_{SFU}$ &    SPEC &    UPOS &           STL \\

\cmidrule(lr){4-4}\cmidrule(lr){5-5}\cmidrule(lr){6-6}\cmidrule(lr){7-7}\cmidrule(lr){8-8}\cmidrule(lr){9-9}\cmidrule(lr){10-10}

\multirow{8}{*}{\rotatebox[origin=c]{90}{MAMS}} & \multirow{2}{*}{acc-s} & GloVe &  80.73$^\star$ \sd{\sd{(0.22)}} &   80.98 \sd{\sd{(1.17)}} &  80.89$^\star$ \sd{\sd{(0.45)}} &   80.68$^\star$ \sd{\sd{(0.48)}} &  80.87$^\star$ \sd{\sd{(0.53)}} &  81.36 \sd{\sd{(0.82)}} &  \textbf{82.14} \sd{\sd{(0.77)}} \\
&      & TL  &  84.37 \sd{\sd{(0.22)}} &   83.99$^\star$ \sd{\sd{(0.25)}} &  \posbox{\textbf{84.73}} \sd{\sd{(0.34)}} &   84.18 \sd{\sd{(0.49)}} &  84.17 \sd{\sd{(0.53)}} &  83.79$^\star$ \sd{\sd{(0.40)}} &  84.43 \sd{\sd{(0.41)}} \\

\cmidrule(lr){4-4}\cmidrule(lr){5-5}\cmidrule(lr){6-6}\cmidrule(lr){7-7}\cmidrule(lr){8-8}\cmidrule(lr){9-9}\cmidrule(lr){10-10}

& \multirow{2}{*}{F1-s} & GloVe &  80.20$^\star$ \sd{\sd{(0.24)}} &   80.48$^\star$ \sd{\sd{(1.15)}} &  80.38$^\star$ \sd{\sd{(0.50)}} &   80.14$^\star$ \sd{\sd{(0.57)}} &  80.38$^\star$ \sd{\sd{(0.56)}} &  80.76 \sd{\sd{(0.95)}} &  \textbf{81.67} \sd{\sd{(0.77)}} \\
&      & TL  &  \posbox{84.14}$^\star$ \sd{\sd{(0.24)}} &   83.72$^\star$ \sd{\sd{(0.21)}} &  \posbox{\textbf{84.46}} \sd{\sd{(0.35)}} &   83.96$^\star$ \sd{\sd{(0.47)}} &  83.93 \sd{\sd{(0.47)}} &  83.54$^\star$ \sd{\sd{(0.42)}} &  84.12 \sd{\sd{(0.45)}} \\

\cmidrule(lr){4-4}\cmidrule(lr){5-5}\cmidrule(lr){6-6}\cmidrule(lr){7-7}\cmidrule(lr){8-8}\cmidrule(lr){9-9}\cmidrule(lr){10-10}

& \multirow{2}{*}{extraction} & GloVe &  78.93 \sd{\sd{(0.66)}} &   79.11 \sd{\sd{(0.52)}} &  \posbox{\textbf{79.24}} \sd{\sd{(0.62)}} &   79.00 \sd{\sd{(0.47)}} &  78.86 \sd{\sd{(0.67)}} &  78.68 \sd{\sd{(0.35)}} &  79.15 \sd{\sd{(0.39)}} \\
&      & TL  &  77.81 \sd{\sd{(0.48)}} &   77.86 \sd{\sd{(0.59)}} &  77.62$^\star$ \sd{\sd{(0.34)}} &   78.32 \sd{\sd{(0.31)}} &  77.59$^\star$ \sd{\sd{(0.21)}} &  \posbox{\textbf{78.54}} \sd{\sd{(0.38)}} &  78.35 \sd{\sd{(0.40)}} \\

\cmidrule(lr){4-4}\cmidrule(lr){5-5}\cmidrule(lr){6-6}\cmidrule(lr){7-7}\cmidrule(lr){8-8}\cmidrule(lr){9-9}\cmidrule(lr){10-10}

& \multirow{2}{*}{targeted} & GloVe &  63.72$^\star$ \sd{\sd{(0.63)}} &   64.06$^\star$ \sd{\sd{(0.66)}} &  64.10$^\star$ \sd{\sd{(0.48)}} &   63.74$^\star$ \sd{\sd{(0.28)}} &  63.76$^\star$ \sd{\sd{(0.21)}} &  64.01$^\star$ \sd{\sd{(0.62)}} &  \textbf{65.01} \sd{\sd{(0.44)}} \\
&      & TL  &  65.65 \sd{\sd{(0.39)}} &   65.39$^\star$ \sd{\sd{(0.47)}} &  65.77$^\star$ \sd{\sd{(0.44)}} &   65.93 \sd{\sd{(0.20)}} &  65.31$^\star$ \sd{\sd{(0.56)}} &  65.81 \sd{\sd{(0.41)}} &  \textbf{66.15} \sd{\sd{(0.54)}} \\[2ex]

\hline\\ [-1.5ex]

\multirow{8}{*}{\rotatebox[origin=c]{90}{Restaurant}} & \multirow{2}{*}{acc-s} & GloVe &  78.42$^\star$ \sd{\sd{(0.78)}} &   \posbox{78.78} \sd{\sd{(0.67)}} &  \posbox{\textbf{79.58}} \sd{\sd{(0.89)}} &   78.75 \sd{\sd{(0.52)}} &  78.31$^\star$ \sd{\sd{(0.78)}} &  \posbox{79.14} \sd{\sd{(0.41)}} &  78.76 \sd{\sd{(0.37)}} \\
&      & TL  &  \posbox{81.90} \sd{\sd{(0.69)}} &   \posbox{\textbf{81.90}} \sd{\sd{(0.69)}} &  81.53 \sd{\sd{(0.86)}} &   \posbox{81.89} \sd{\sd{(0.84)}} &  81.02 \sd{\sd{(0.85)}} &  80.47$^\star$ \sd{\sd{(1.10)}} &  81.77 \sd{\sd{(0.46)}} \\

\cmidrule(lr){4-4}\cmidrule(lr){5-5}\cmidrule(lr){6-6}\cmidrule(lr){7-7}\cmidrule(lr){8-8}\cmidrule(lr){9-9}\cmidrule(lr){10-10}

& \multirow{2}{*}{F1-s} & GloVe &  62.89 \sd{\sd{(2.84)}} &   \posbox{64.01} \sd{\sd{(2.56)}} &  \posbox{\textbf{65.37}} \sd{\sd{(1.47)}} &   62.49$^\star$ \sd{\sd{(0.86)}} &  62.54$^\star$ \sd{\sd{(2.28)}} &  63.00$^\star$ \sd{\sd{(1.64)}} &  63.15$^\star$ \sd{\sd{(1.94)}} \\
&      & TL  &  67.98 \sd{\sd{(3.46)}} &   69.26 \sd{\sd{(1.07)}} &  69.09 \sd{\sd{(1.72)}} &   68.18 \sd{\sd{(1.90)}} &  67.54 \sd{\sd{(3.88)}} &  67.14$^\star$ \sd{\sd{(1.05)}} &  \textbf{69.37} \sd{\sd{(0.97)}} \\

\cmidrule(lr){4-4}\cmidrule(lr){5-5}\cmidrule(lr){6-6}\cmidrule(lr){7-7}\cmidrule(lr){8-8}\cmidrule(lr){9-9}\cmidrule(lr){10-10}

& \multirow{2}{*}{extraction} & GloVe &  78.22$^\star$ \sd{\sd{(0.78)}} &   \posbox{79.20} \sd{\sd{(1.07)}} &  78.85 \sd{\sd{(1.08)}} &   78.38$^\star$ \sd{\sd{(0.73)}} &  78.21 \sd{\sd{(1.37)}} &  \posbox{\textbf{79.62}} \sd{\sd{(0.48)}} &  79.18 \sd{\sd{(0.76)}} \\
&      & TL  &  81.69 \sd{\sd{(0.71)}} &   81.84 \sd{\sd{(0.88)}} &  \textbf{\posbox{82.56}} \sd{\sd{(0.79)}} &   82.25 \sd{\sd{(0.22)}} &  82.07 \sd{\sd{(0.68)}} &  \posbox{82.48} \sd{\sd{(0.61)}} &  82.33 \sd{\sd{(0.52)}} \\

\cmidrule(lr){4-4}\cmidrule(lr){5-5}\cmidrule(lr){6-6}\cmidrule(lr){7-7}\cmidrule(lr){8-8}\cmidrule(lr){9-9}\cmidrule(lr){10-10}

& \multirow{2}{*}{targeted} & GloVe &  61.34$^\star$ \sd{\sd{(0.73)}} &   \posbox{62.39} \sd{\sd{(0.95)}} &  \posbox{62.75} \sd{\sd{(1.26)}} &   61.73$^\star$ \sd{\sd{(0.77)}} &  61.25$^\star$ \sd{\sd{(1.26)}} &  \posbox{\textbf{63.02}} \sd{\sd{(0.42)}} &  62.36 \sd{\sd{(0.60)}} \\
&      & TL  &  66.90 \sd{\sd{(0.40)}} &   67.03 \sd{\sd{(1.06)}} &  67.31 \sd{\sd{(0.32)}} &   \posbox{\textbf{67.36}} \sd{\sd{(0.66)}} &  66.49 \sd{\sd{(0.52)}} &  66.38$^\star$ \sd{\sd{(1.31)}} &  67.32 \sd{\sd{(0.55)}} \\

\bottomrule
\end{tabular}

    \caption{\accs, \Fs, extraction (\Fa) and full targeted (\Fi) results for MAMS and Restaurant \textbf{development split}, where the values represent the mean (standard deviation) of five runs with a different random seed. The \textbf{bold} values represent the best model, while \posbox{highlighted} models are those that perform better than the single task baseline. The $^\star$ represent the models that are statistically significantly worse than the best performing model on the respective dataset, metric and embedding at a 95\% confidence level.}
    \label{tab:Main_Development_Results_All_Metrics_Restaurant_MAMS}
\end{table*}

\clearpage
\section{Additional Negation and Speculation Result Tables}
\label{appendix_neg_spec_results}

\begin{table*}[!h]
    \centering
    \begin{tabular}{ccP{1.2cm}P{1.2cm}P{1.2cm}P{1.2cm}P{1.2cm}P{1.3cm}P{1.2cm}}
\toprule
   &  &      NEG$_{CD}$ &      DR &     LEX &     NEG$_{SFU}$ &    SPEC &    UPOS &           STL \\

\cmidrule(lr){3-3}\cmidrule(lr){4-4}\cmidrule(lr){5-5}\cmidrule(lr){6-6}\cmidrule(lr){7-7}\cmidrule(lr){8-8}\cmidrule(lr){9-9}
\multicolumn{2}{c}{} & \multicolumn{7}{c}{Laptop$_{Neg}$} \\
\multirow{2}{*}{F1-s} & GloVe &  40.88 \sd{\sd{(2.17)}} &  38.11$^\star$ \sd{\sd{(1.71)}} &  38.89 \sd{\sd{(3.32)}} &  \posbox{\textbf{42.33}} \sd{\sd{(1.19)}} &  39.46$^\star$ \sd{\sd{(2.98)}} &  38.11$^\star$ \sd{\sd{(2.02)}} &  37.13$^\star$ \sd{\sd{(2.78)}} \\
& TL  &  44.81 \sd{\sd{(2.40)}} &  45.05 \sd{\sd{(3.47)}} &  44.58 \sd{\sd{(2.29)}} &  43.53 \sd{\sd{(2.74)}} &  43.83 \sd{\sd{(1.90)}} &  44.77 \sd{\sd{(1.54)}} &  \textbf{45.08} \sd{\sd{(2.68)}} \\[2ex]

\hline\\ [-1.5ex]

\multicolumn{2}{c}{} & \multicolumn{7}{c}{Restaurant$_{Neg}$} \\
\multirow{2}{*}{F1-s} & GloVe &  46.58 \sd{\sd{(3.24)}} &  44.16$^\star$ \sd{\sd{(2.18)}} &  42.74$^\star$ \sd{\sd{(1.09)}} &  \posbox{\textbf{47.65}} \sd{\sd{(1.35)}} &  44.00 \sd{\sd{(3.99)}} &  44.78 \sd{\sd{(1.85)}} &  44.14$^\star$ \sd{\sd{(1.89)}} \\
& TL  &  52.85$^\star$ \sd{\sd{(1.69)}} &  54.41 \sd{\sd{(1.51)}} &  54.08 \sd{\sd{(3.87)}} &  52.54$^\star$ \sd{\sd{(1.99)}} &  \posbox{\textbf{55.63}} \sd{\sd{(1.65)}} &  52.16$^\star$ \sd{\sd{(2.01)}} &  53.59$^\star$ \sd{\sd{(1.95)}} \\

\bottomrule
\end{tabular}
    \caption{\Fs results for the \textbf{negation test split}, where the values represent the mean (standard deviation) of five runs with a different random seed. The \textbf{bold} values represent the best model, while \posbox{highlighted} models are those that perform better than the single task baseline. The $^\star$ represent the models that are statistically significantly worse than the best performing model on the respective dataset, metric and embedding at a 95\% confidence level.}
    \label{tab:Negation_Only_Test_F1_Scores}
\end{table*}

\begin{table*}[!h]
    \centering
    \begin{tabular}{ccP{1.2cm}P{1.2cm}P{1.2cm}P{1.2cm}P{1.2cm}P{1.3cm}P{1.2cm}}
\toprule
   &  &      NEG$_{CD}$ &      DR &     LEX &     NEG$_{SFU}$ &    SPEC &    UPOS &           STL \\

\cmidrule(lr){3-3}\cmidrule(lr){4-4}\cmidrule(lr){5-5}\cmidrule(lr){6-6}\cmidrule(lr){7-7}\cmidrule(lr){8-8}\cmidrule(lr){9-9}
\multicolumn{2}{c}{} & \multicolumn{7}{c}{Laptop$_{Spec}$} \\

\multirow{2}{*}{F1-s} & GloVe &  \posbox{32.74} \sd{\sd{(2.35)}} &  \posbox{33.14} \sd{\sd{(0.98)}} &  \textbf{\posbox{35.24}} \sd{\sd{(2.16)}} &  \posbox{33.62} \sd{\sd{(2.59)}} &  \posbox{33.83} \sd{\sd{(1.62)}} &  \posbox{33.21} \sd{\sd{(1.00)}} &  31.99$^\star$ \sd{\sd{(1.92)}} \\
& TL  &  33.02 \sd{\sd{(4.07)}} &  33.33 \sd{\sd{(1.56)}} &  33.14 \sd{\sd{(2.10)}} &  31.72$^\star$ \sd{\sd{(0.88)}} &  33.25 \sd{\sd{(2.14)}} &  32.71 \sd{\sd{(1.38)}} &  \textbf{34.08} \sd{\sd{(1.40)}} \\[2ex]

\hline\\ [-1.5ex]

\multicolumn{2}{c}{} & \multicolumn{7}{c}{Restaurant$_{Spec}$} \\

\multirow{2}{*}{F1-s} & GloVe &  55.27 \sd{\sd{(3.82)}} &  \posbox{\textbf{57.77}} \sd{\sd{(2.91)}} &  56.27$^\star$ \sd{\sd{(2.36)}} &  55.59$^\star$ \sd{\sd{(1.09)}} &  \posbox{57.35} \sd{\sd{(3.75)}} &  56.55 \sd{\sd{(3.14)}} &  57.32 \sd{\sd{(2.30)}} \\
& TL  &  58.84$^\star$ \sd{\sd{(1.58)}} &  \posbox{60.95} \sd{\sd{(0.98)}} &  \posbox{\textbf{62.36}} \sd{\sd{(2.86)}} &  58.44$^\star$ \sd{\sd{(2.24)}} &  60.52 \sd{\sd{(3.25)}} &  59.23 \sd{\sd{(1.81)}} &  60.74 \sd{\sd{(2.32)}} \\

\bottomrule
\end{tabular}
    \caption{\Fs results for the \textbf{speculation test split}, where the values represent the mean (standard deviation) of five runs with a different random seed. The \textbf{bold} values represent the best model, while \posbox{highlighted} models are those that perform better than the single task baseline. The $^\star$ represent the models that are statistically significantly worse than the best performing model on the respective dataset, metric and embedding at a 95\% confidence level.}
    \label{tab:Speculation_Only_Test_F1_Scores}
\end{table*}

\begin{table*}[!h]
    \centering
    \begin{tabular}{cccP{1.2cm}P{1.2cm}P{1.2cm}P{1.2cm}P{1.2cm}P{1.3cm}P{1.2cm}}
\toprule
 &   &  &      NEG$_{CD}$ &      DR &     LEX &     NEG$_{SFU}$ &    SPEC &    UPOS &           STL \\

\cmidrule(lr){4-4}\cmidrule(lr){5-5}\cmidrule(lr){6-6}\cmidrule(lr){7-7}\cmidrule(lr){8-8}\cmidrule(lr){9-9}\cmidrule(lr){10-10}

\multirow{8}{*}{\rotatebox[origin=c]{90}{Laptop$_{Neg}$}} & \multirow{2}{*}{acc-s} & GloVe &  \posbox{\textbf{37.07}} \sd{\sd{(3.78)}} &  \posbox{32.99}$^\star$ \sd{\sd{(0.96)}} &  \posbox{33.98} \sd{\sd{(2.60)}} &  \posbox{36.64} \sd{\sd{(3.17)}} &  \posbox{34.83} \sd{\sd{(1.09)}} &  \posbox{32.09}$^\star$ \sd{\sd{(3.41)}} &  29.45$^\star$ \sd{\sd{(1.46)}} \\
&      & TL  &  \posbox{41.84}$^\star$ \sd{\sd{(0.54)}} &  39.38$^\star$ \sd{\sd{(2.69)}} &  40.85$^\star$ \sd{\sd{(2.50)}} &  \posbox{\textbf{45.40}} \sd{\sd{(1.70)}} &  \posbox{43.00}$^\star$ \sd{\sd{(1.57)}} &  38.77$^\star$ \sd{\sd{(2.64)}} &  41.15$^\star$ \sd{\sd{(3.28)}} \\

\cmidrule(lr){4-4}\cmidrule(lr){5-5}\cmidrule(lr){6-6}\cmidrule(lr){7-7}\cmidrule(lr){8-8}\cmidrule(lr){9-9}\cmidrule(lr){10-10}

& \multirow{2}{*}{F1-s} & GloVe &  \posbox{34.02} \sd{\sd{(2.05)}} &  \posbox{31.48}$^\star$ \sd{\sd{(2.66)}} &  \posbox{33.06} \sd{\sd{(2.21)}} &  \textbf{\posbox{34.61}} \sd{\sd{(3.34)}} &  \posbox{33.66} \sd{\sd{(1.13)}} &  \posbox{28.92}$^\star$ \sd{\sd{(4.11)}} &  24.92$^\star$ \sd{\sd{(1.33)}} \\
&      & TL  &  38.26$^\star$ \sd{\sd{(1.67)}} &  38.05 \sd{\sd{(3.06)}} &  \posbox{39.42} \sd{\sd{(4.02)}} &  \textbf{\posbox{42.32}} \sd{\sd{(3.45)}} &  \posbox{41.39} \sd{\sd{(1.97)}} &  36.73 \sd{\sd{(3.67)}} &  38.92 \sd{\sd{(3.45)}} \\

\cmidrule(lr){4-4}\cmidrule(lr){5-5}\cmidrule(lr){6-6}\cmidrule(lr){7-7}\cmidrule(lr){8-8}\cmidrule(lr){9-9}\cmidrule(lr){10-10}

& \multirow{2}{*}{extraction} & GloVe &  72.49 \sd{\sd{(1.45)}} &  74.40 \sd{\sd{(1.64)}} &  73.91 \sd{\sd{(0.99)}} &  71.25$^\star$ \sd{\sd{(1.20)}} &  73.83 \sd{\sd{(0.93)}} &  73.33 \sd{\sd{(1.90)}} &  \textbf{74.60} \sd{\sd{(1.51)}} \\
&      & TL  &  80.94 \sd{\sd{(1.64)}} &  81.31 \sd{\sd{(1.29)}} &  81.21 \sd{\sd{(1.36)}} &  81.94 \sd{\sd{(1.46)}} &  79.00$^\star$ \sd{\sd{(0.87)}} &  81.92 \sd{\sd{(1.36)}} &  \textbf{82.75} \sd{\sd{(1.80)}} \\

\cmidrule(lr){4-4}\cmidrule(lr){5-5}\cmidrule(lr){6-6}\cmidrule(lr){7-7}\cmidrule(lr){8-8}\cmidrule(lr){9-9}\cmidrule(lr){10-10}

& \multirow{2}{*}{targeted} & GloVe &  \posbox{\textbf{26.87}} \sd{\sd{(2.75)}} &  \posbox{24.56} \sd{\sd{(1.21)}} &  \posbox{25.12} \sd{\sd{(1.99)}} &  \posbox{26.14} \sd{\sd{(2.55)}} &  \posbox{25.71} \sd{\sd{(0.68)}} &  \posbox{23.54} \sd{\sd{(2.66)}} &  21.98$^\star$ \sd{\sd{(1.26)}} \\
&      & TL  &  33.86$^\star$ \sd{\sd{(0.71)}} &  32.04$^\star$ \sd{\sd{(2.54)}} &  33.14$^\star$ \sd{\sd{(1.56)}} &  \posbox{\textbf{37.20}} \sd{\sd{(1.59)}} &  33.98$^\star$ \sd{\sd{(1.47)}} &  31.78$^\star$ \sd{\sd{(2.46)}} &  34.01$^\star$ \sd{\sd{(2.29)}} \\[2ex]

\hline\\ [-1.5ex]

\multirow{8}{*}{\rotatebox[origin=c]{90}{Restaurant$_{Neg}$}} & \multirow{2}{*}{acc-s} & GloVe &  \posbox{46.02} \sd{\sd{(4.88)}} &  \posbox{43.13}$^\star$ \sd{\sd{(3.24)}} &  41.06$^\star$ \sd{\sd{(3.36)}} &  \posbox{\textbf{49.02}} \sd{\sd{(1.31)}} &  41.65$^\star$ \sd{\sd{(4.06)}} &  \posbox{44.02}$^\star$ \sd{\sd{(3.09)}} &  42.69$^\star$ \sd{\sd{(2.01)}} \\
&      & TL  &  \posbox{53.79} \sd{\sd{(2.56)}} &  \posbox{54.40} \sd{\sd{(3.63)}} &  52.25 \sd{\sd{(3.63)}} &  \posbox{\textbf{54.42}} \sd{\sd{(3.18)}} &  \posbox{53.16} \sd{\sd{(1.92)}} &  \posbox{54.31} \sd{\sd{(2.49)}} &  52.25 \sd{\sd{(2.51)}} \\

\cmidrule(lr){4-4}\cmidrule(lr){5-5}\cmidrule(lr){6-6}\cmidrule(lr){7-7}\cmidrule(lr){8-8}\cmidrule(lr){9-9}\cmidrule(lr){10-10}

& \multirow{2}{*}{F1-s} & GloVe &  \posbox{40.03} \sd{\sd{(5.30)}} &  \posbox{38.56}$^\star$ \sd{\sd{(3.10)}} &  37.54 \sd{\sd{(3.69)}} &  \posbox{\textbf{41.05}} \sd{\sd{(2.45)}} &  37.01 \sd{\sd{(4.16)}} &  38.03 \sd{\sd{(3.40)}} &  38.45 \sd{\sd{(2.28)}} \\
&      & TL  &  48.03 \sd{\sd{(3.48)}} &  \posbox{49.13} \sd{\sd{(2.62)}} &  \posbox{48.31} \sd{\sd{(3.71)}} &  \posbox{49.18} \sd{\sd{(3.58)}} &  \posbox{48.80} \sd{\sd{(2.03)}} &  \posbox{\textbf{49.42}} \sd{\sd{(2.76)}} &  48.06 \sd{\sd{(2.12)}} \\

\cmidrule(lr){4-4}\cmidrule(lr){5-5}\cmidrule(lr){6-6}\cmidrule(lr){7-7}\cmidrule(lr){8-8}\cmidrule(lr){9-9}\cmidrule(lr){10-10}

& \multirow{2}{*}{extraction} & GloVe &  \posbox{81.74} \sd{\sd{(0.77)}} &  \posbox{82.37} \sd{\sd{(0.64)}} &  \posbox{82.36} \sd{\sd{(0.80)}} &  80.61$^\star$ \sd{\sd{(0.72)}} &  \posbox{81.34} \sd{\sd{(1.48)}} &  \posbox{\textbf{82.38}} \sd{\sd{(1.00)}} &  81.32 \sd{\sd{(0.37)}} \\
&      & TL  &  84.08 \sd{\sd{(0.72)}} &  82.87$^\star$ \sd{\sd{(0.66)}} &  84.32 \sd{\sd{(0.68)}} &  83.71 \sd{\sd{(0.55)}} &  83.45 \sd{\sd{(1.09)}} &  84.02 \sd{\sd{(1.12)}} &  \textbf{84.84} \sd{\sd{(0.99)}} \\

\cmidrule(lr){4-4}\cmidrule(lr){5-5}\cmidrule(lr){6-6}\cmidrule(lr){7-7}\cmidrule(lr){8-8}\cmidrule(lr){9-9}\cmidrule(lr){10-10}

& \multirow{2}{*}{targeted} & GloVe &  \posbox{37.61} \sd{\sd{(3.86)}} &  \posbox{35.54}$^\star$ \sd{\sd{(2.81)}} &  33.82$^\star$ \sd{\sd{(2.78)}} &  \posbox{\textbf{39.51}} \sd{\sd{(1.03)}} &  33.88$^\star$ \sd{\sd{(3.38)}} &  \posbox{36.27}$^\star$ \sd{\sd{(2.64)}} &  34.71$^\star$ \sd{\sd{(1.58)}} \\
&      & TL  &  45.23 \sd{\sd{(2.19)}} &  \posbox{45.07} \sd{\sd{(2.90)}} &  44.04 \sd{\sd{(2.77)}} &  \posbox{45.57} \sd{\sd{(2.91)}} &  \posbox{44.38} \sd{\sd{(2.05)}} &  \posbox{\textbf{45.62}} \sd{\sd{(1.88)}} &  44.31 \sd{\sd{(1.74)}} \\

\bottomrule
\end{tabular}
    \caption{\accs, \Fs, extraction (\Fa) and full targeted (\Fi) results for the \textbf{negation development split}, where the values represent the mean (standard deviation) of five runs with a different random seed. The \textbf{bold} values represent the best model, while \posbox{highlighted} models are those that perform better than the single task baseline. The $^\star$ represent the models that are statistically significantly worse than the best performing model on the respective dataset, metric and embedding at a 95\% confidence level.}
    \label{tab:Negation_Only_Validation_Results}
\end{table*}

\begin{table*}[!h]
    \centering
    \begin{tabular}{cccP{1.2cm}P{1.2cm}P{1.2cm}P{1.2cm}P{1.2cm}P{1.3cm}P{1.2cm}}
\toprule
 &   &  &      NEG$_{CD}$ &      DR &     LEX &     NEG$_{SFU}$ &    SPEC &    UPOS &           STL \\

\cmidrule(lr){4-4}\cmidrule(lr){5-5}\cmidrule(lr){6-6}\cmidrule(lr){7-7}\cmidrule(lr){8-8}\cmidrule(lr){9-9}\cmidrule(lr){10-10}

\multirow{8}{*}{\rotatebox[origin=c]{90}{Laptop$_{Spec}$}} & \multirow{2}{*}{acc-s} & GloVe &  \posbox{33.56}$^\star$ \sd{\sd{(2.39)}} &  \posbox{35.00} \sd{\sd{(2.94)}} &  31.70$^\star$ \sd{\sd{(1.64)}} &  \posbox{34.67} \sd{\sd{(2.21)}} &  \posbox{\textbf{37.24}} \sd{\sd{(2.48)}} &  31.59$^\star$ \sd{\sd{(1.57)}} &  32.57$^\star$ \sd{\sd{(1.94)}} \\
&      & TL  &  \posbox{34.73} \sd{\sd{(1.94)}} &  \posbox{35.02} \sd{\sd{(3.50)}} &  \posbox{\textbf{35.16}} \sd{\sd{(2.48)}} &  32.79 \sd{\sd{(1.43)}} &  34.24 \sd{\sd{(1.95)}} &  34.27 \sd{\sd{(1.56)}} &  34.28 \sd{\sd{(1.19)}} \\

\cmidrule(lr){4-4}\cmidrule(lr){5-5}\cmidrule(lr){6-6}\cmidrule(lr){7-7}\cmidrule(lr){8-8}\cmidrule(lr){9-9}\cmidrule(lr){10-10}

& \multirow{2}{*}{F1-s} & GloVe &  \posbox{32.66} \sd{\sd{(2.33)}} &  \posbox{33.99} \sd{\sd{(3.14)}} &  30.72$^\star$ \sd{\sd{(1.33)}} &  \posbox{33.20} \sd{\sd{(2.36)}} &  \posbox{\textbf{34.99}} \sd{\sd{(2.91)}} &  29.88$^\star$ \sd{\sd{(1.50)}} &  30.75 \sd{\sd{(2.02)}} \\           
&      & TL  &  \posbox{33.60} \sd{\sd{(2.30)}} &  \posbox{\textbf{34.14}} \sd{\sd{(3.47)}} &  \posbox{33.83} \sd{\sd{(2.83)}} &  30.92 \sd{\sd{(1.48)}} &  \posbox{33.36} \sd{\sd{(1.98)}} &  32.45 \sd{\sd{(1.32)}} &  32.47 \sd{\sd{(1.67)}} \\

\cmidrule(lr){4-4}\cmidrule(lr){5-5}\cmidrule(lr){6-6}\cmidrule(lr){7-7}\cmidrule(lr){8-8}\cmidrule(lr){9-9}\cmidrule(lr){10-10}

& \multirow{2}{*}{extraction} & GloVe &  71.10 \sd{\sd{(1.97)}} &  69.42 \sd{\sd{(2.25)}} &  72.61 \sd{\sd{(0.95)}} &  70.46 \sd{\sd{(1.41)}} &  69.33$^\star$ \sd{\sd{(1.58)}} &  70.58$^\star$ \sd{\sd{(2.46)}} &  \textbf{73.04} \sd{\sd{(2.48)}} \\
&      & TL  &  80.50 \sd{\sd{(0.95)}} &  80.25 \sd{\sd{(0.90)}} &  79.91$^\star$ \sd{\sd{(0.61)}} &  80.83 \sd{\sd{(0.61)}} &  79.35$^\star$ \sd{\sd{(1.34)}} &  80.95 \sd{\sd{(0.43)}} &  \textbf{82.00} \sd{\sd{(1.32)}} \\

\cmidrule(lr){4-4}\cmidrule(lr){5-5}\cmidrule(lr){6-6}\cmidrule(lr){7-7}\cmidrule(lr){8-8}\cmidrule(lr){9-9}\cmidrule(lr){10-10}

& \multirow{2}{*}{targeted} & GloVe &  \posbox{23.88} \sd{\sd{(2.05)}} &  \posbox{24.26} \sd{\sd{(1.81)}} &  23.02$^\star$ \sd{\sd{(1.35)}} &  \posbox{24.44}$^\star$ \sd{\sd{(1.81)}} &  \posbox{\textbf{25.82}} \sd{\sd{(1.82)}} &  22.32$^\star$ \sd{\sd{(1.73)}} &  23.80 \sd{\sd{(1.77)}} \\
&      & TL  &  27.96 \sd{\sd{(1.60)}} &  28.11 \sd{\sd{(2.92)}} &  28.08 \sd{\sd{(1.81)}} &  26.51 \sd{\sd{(1.34)}} &  27.16 \sd{\sd{(1.39)}} &  27.74 \sd{\sd{(1.31)}} &  \textbf{28.12} \sd{\sd{(1.32)}} \\[2ex]

\hline\\ [-1.5ex]

\multirow{8}{*}{\rotatebox[origin=c]{90}{Restaurant$_{Spec}$}} & \multirow{2}{*}{acc-s} & GloVe &  35.54$^\star$ \sd{\sd{(0.90)}} &  \posbox{\textbf{40.09}} \sd{\sd{(2.95)}} &  \posbox{37.98} \sd{\sd{(2.38)}} &  37.18 \sd{\sd{(2.14)}} &  \posbox{37.97} \sd{\sd{(2.28)}} &  \posbox{38.32} \sd{\sd{(1.31)}} &  37.23 \sd{\sd{(1.65)}} \\
&      & TL  &  38.80 \sd{\sd{(2.17)}} &  38.50 \sd{\sd{(1.19)}} &  \posbox{40.72} \sd{\sd{(1.18)}} &  \posbox{\textbf{40.84}} \sd{\sd{(2.30)}} &  39.69 \sd{\sd{(1.69)}} &  39.49 \sd{\sd{(0.89)}} &  40.55 \sd{\sd{(1.16)}} \\

\cmidrule(lr){4-4}\cmidrule(lr){5-5}\cmidrule(lr){6-6}\cmidrule(lr){7-7}\cmidrule(lr){8-8}\cmidrule(lr){9-9}\cmidrule(lr){10-10}

& \multirow{2}{*}{F1-s} & GloVe &  31.46$^\star$ \sd{\sd{(1.47)}} &  \textbf{\posbox{35.99}} \sd{\sd{(3.98)}} &  \posbox{34.37} \sd{\sd{(2.89)}} &  32.53 \sd{\sd{(1.82)}} &  \posbox{34.02} \sd{\sd{(2.03)}} &  32.92 \sd{\sd{(1.74)}} &  33.18 \sd{\sd{(1.15)}} \\
&      & TL  &  33.00$^\star$ \sd{\sd{(3.34)}} &  33.28 \sd{\sd{(1.65)}} &  \posbox{\textbf{35.92}} \sd{\sd{(1.84)}} &  35.26 \sd{\sd{(3.37)}} &  34.47 \sd{\sd{(3.78)}} &  35.08 \sd{\sd{(1.78)}} &  35.40 \sd{\sd{(1.59)}} \\

\cmidrule(lr){4-4}\cmidrule(lr){5-5}\cmidrule(lr){6-6}\cmidrule(lr){7-7}\cmidrule(lr){8-8}\cmidrule(lr){9-9}\cmidrule(lr){10-10}

& \multirow{2}{*}{extraction} & GloVe &  78.26$^\star$ \sd{\sd{(0.92)}} &  80.38 \sd{\sd{(1.86)}} &  80.33 \sd{\sd{(1.13)}} &  79.75$^\star$ \sd{\sd{(1.05)}} &  79.99$^\star$ \sd{\sd{(1.20)}} &  \posbox{\textbf{81.98}} \sd{\sd{(1.04)}} &  80.59 \sd{\sd{(1.29)}} \\
&      & TL  &  83.91 \sd{\sd{(0.67)}} &  83.93 \sd{\sd{(1.03)}} &  84.60 \sd{\sd{(0.63)}} &  84.77 \sd{\sd{(0.29)}} &  83.40 \sd{\sd{(1.58)}} &  \posbox{\textbf{85.03}} \sd{\sd{(1.47)}} &  84.85 \sd{\sd{(0.83)}} \\

\cmidrule(lr){4-4}\cmidrule(lr){5-5}\cmidrule(lr){6-6}\cmidrule(lr){7-7}\cmidrule(lr){8-8}\cmidrule(lr){9-9}\cmidrule(lr){10-10}

& \multirow{2}{*}{targeted} & GloVe &  27.82$^\star$ \sd{\sd{(0.79)}} &  \posbox{\textbf{32.25}} \sd{\sd{(2.79)}} &  \posbox{30.51}$^\star$ \sd{\sd{(2.06)}} &  29.64$^\star$ \sd{\sd{(1.60)}} &  \posbox{30.37} \sd{\sd{(1.91)}} &  \posbox{31.43} \sd{\sd{(1.42)}} &  30.02 \sd{\sd{(1.74)}} \\
&      & TL  &  32.56 \sd{\sd{(1.83)}} &  32.31$^\star$ \sd{\sd{(1.13)}} &  \posbox{34.45} \sd{\sd{(1.00)}} &  \posbox{\textbf{34.62}} \sd{\sd{(1.96)}} &  33.08 \sd{\sd{(1.01)}} &  33.59 \sd{\sd{(1.20)}} &  34.41 \sd{\sd{(1.23)}} \\

\bottomrule
\end{tabular}
    \caption{\accs, \Fs, extraction (\Fa) and full targeted (\Fi) results for the \textbf{speculation development split}, where the values represent the mean (standard deviation) of five runs with a different random seed. The \textbf{bold} values represent the best model, while \posbox{highlighted} models are those that perform better than the single task baseline. The $^\star$ represent the models that are statistically significantly worse than the best performing model on the respective dataset, metric and embedding at a 95\% confidence level.}
    \label{tab:Speculation_Only_Validation_Results}
\end{table*}

\clearpage 
\section{Hyperparameter Search Space}
\label{appendix:hyperparameter_search}

\begin{table*}[!h]
    \centering
    \begin{tabular}{cc}
\hline
    \textbf{GPU Infrastructure} & 1 GeForce GTX 1060 6GB GPU \\
\hline
    \textbf{CPU Infrastructure} & AMD Ryzen 5 1600 CPU \\
\hline
    \textbf{Number of search trials} & 30 \\
\hline
    \textbf{Search strategy} & uniform sampling \\
\hline
    \textbf{Best validation span F1/F1-i} & 0.6156 \\
\hline
    \textbf{Training duration} & 14232 sec \\
\hline
    \textbf{Model implementation} & \url{https://bit.ly/3lAz6yf} \\
\hline
\vspace{0.05cm}
\end{tabular}
\begin{tabular}{P{0.3\linewidth}P{0.3\linewidth}P{0.3\linewidth}}
\hline
    \textbf{Hyperparameter} & \textbf{Search space} & \textbf{Best assignment} \\
\hline
    embedding & GloVe 300D & GloVe 300D \\
\hline
    embedding trainable & False & False \\
\hline
    number of epochs & 150 & 150 \\
\hline
    patience &  10 &  10 \\
\hline
    metric early stopping monitored & Span F1/F1-i & Span F1/F1-i \\
\hline
    batch size & 32 & 32 \\
\hline
    dropout & \textit{uniform-float}[0, 0.5] & 0.5 \\
\hline
    1\textsuperscript{st} layer LSTM hidden dimension & \textit{uniform-integer}[30, 110] & 60 \\
\hline
    main task LSTM hidden dimension & 50 & 50 \\
\hline
    skip connection between embedding and main task layer & True & True \\
\hline
    learning rate optimiser & Adam & Adam \\
\hline
    learning rate & \textit{loguniform-float}[1e-4, 1e-2] & 1.5e-3 \\
\hline    
    gradient norm & 5.0 & 5.0 \\ 
\hline
    regularisation type & L2 & L2 \\
\hline
    regularisation value & 1e-4 & 1e-4 \\
\hline
\end{tabular}
    \caption{STL search space and best assignment using the Laptop dataset.}
    \label{table:Parameters_STL}
\end{table*}

\begin{table*}[!h]
    \centering
    \begin{tabular}{cc}
\hline
    \textbf{GPU Infrastructure} & 1 GeForce GTX 1060 6GB GPU \\
\hline
    \textbf{CPU Infrastructure} & AMD Ryzen 5 1600 CPU \\
\hline
    \textbf{Number of search trials} & 30 \\
\hline
    \textbf{Search strategy} & uniform sampling \\
\hline
    \textbf{Best validation span F1/F1-i} & 0.6017 \\
\hline
    \textbf{Training duration} & 18473 sec \\
\hline
    \textbf{Model implementation} & \url{https://bit.ly/3lAz6yf} \\
\hline
\vspace{0.05cm}
\end{tabular}
\begin{tabular}{P{0.3\linewidth}P{0.3\linewidth}P{0.3\linewidth}}
\hline
    \textbf{Hyperparameter} & \textbf{Search space} & \textbf{Best assignment} \\
\hline
    embedding & GloVe 300D & GloVe 300D \\
\hline
    embedding trainable & False & False \\
\hline
    number of epochs & 150 & 150 \\
\hline
    patience &  10 &  10 \\
\hline
    metric early stopping monitored & Span F1/F1-i & Span F1/F1-i \\
\hline
    batch size & 32 & 32 \\
\hline
    dropout & \textit{uniform-float}[0, 0.5] & 0.27 \\
\hline
    Shared/1\textsuperscript{st} layer LSTM hidden dimension & \textit{uniform-integer}[30, 110] & 65 \\
\hline
    main task LSTM hidden dimension & 50 & 50 \\
\hline
    skip connection between embedding and main task layer & True & True \\
\hline
    learning rate optimiser & Adam & Adam \\
\hline
    learning rate & \textit{loguniform-float}[1e-4, 1e-2] & 1.9e-3 \\
\hline    
    gradient norm & 5.0 & 5.0 \\ 
\hline
    regularisation type & L2 & L2 \\
\hline
    regularisation value & 1e-4 & 1e-4 \\
\hline
\end{tabular}
    \caption{MTL search space and best assignment using the Laptop dataset. The auxiliary task was negation detection using the Conan Doyle (NEG$_{CD}$) dataset.}
    \label{table:Parameters_MTL}
\end{table*}

\begin{figure*}[!h]
    \centering
    \includegraphics[scale=0.4]{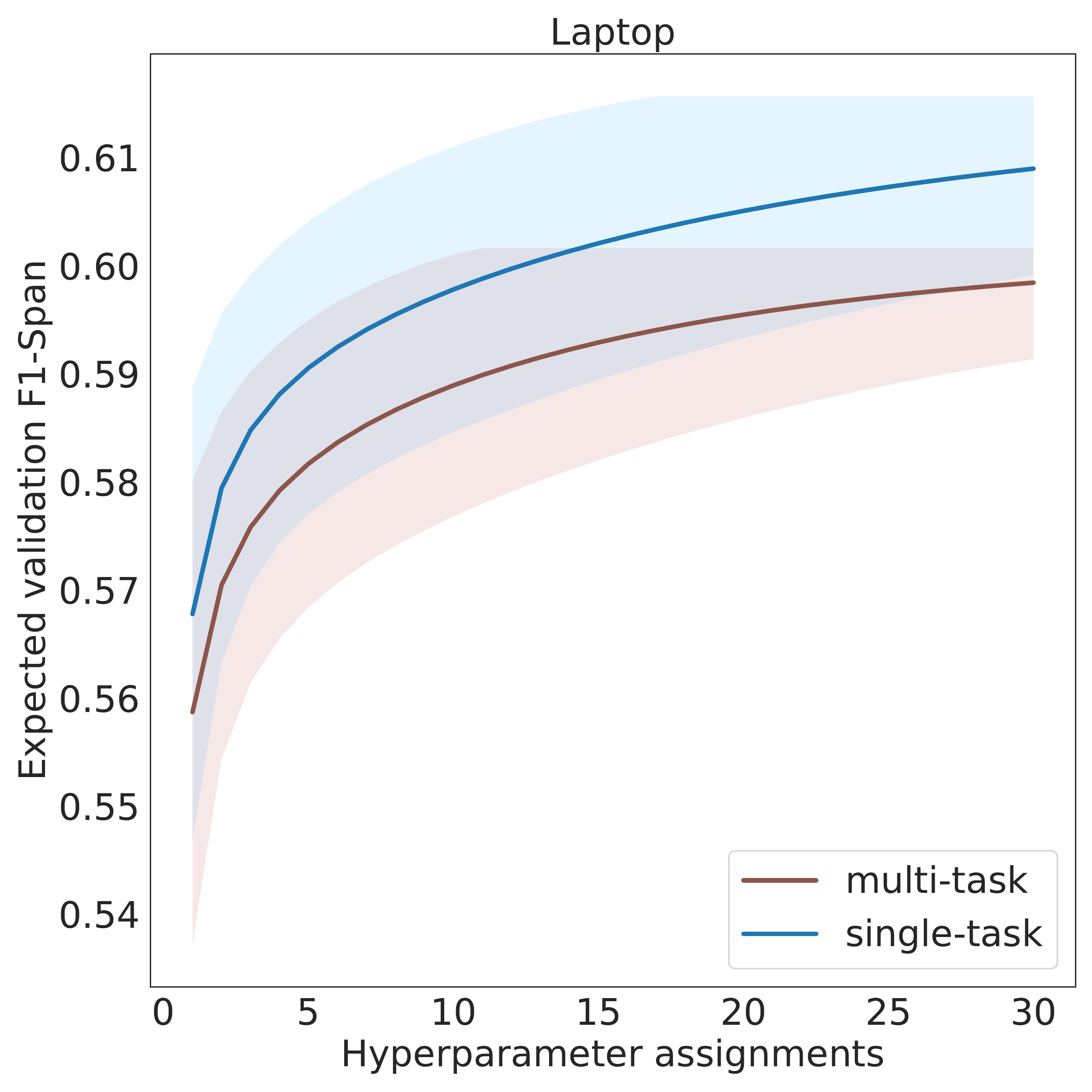}
    \caption{Hyperparameter budget against expected span F1/\Fi performance for the STL and MTL models. The hyperparameter search space is stated within Tables \ref{table:Parameters_STL} and \ref{table:Parameters_MTL} for the STL and MTL models respectively. The shaded areas represent the expected performance $\pm 1$ standard deviation. Note the shaded area does not go beyond the maximum observed validation score as recommended \protect\newcite{dodge-etal-2019-show}.}
    \label{fig:single_task_tuning_laptop_performance}
\end{figure*}

\clearpage 
\section{Additional Reproducibility Statistics}
\label{appendix:additional_reproducibility_statistics}
\begin{table*}[!h]
    \centering
    \begin{tabular}{|c|c|c|c|c|c|}
\hline
    \multirow{4}{*}{\rotatebox[origin=c]{90}{Embedding}} & \multirow{4}{*}{Model} & \multicolumn{4}{c|}{Number of Parameters} \\
\cline{3-6}
    &   & \multicolumn{4}{c|}{Including Auxiliary Task}  \\
\cline{3-6}
    &   & \multicolumn{2}{c|}{Yes} & \multicolumn{2}{c|}{No} \\
\cline{3-6}
    &   & Trainable & All & Trainable & All \\
\hline
    \multirow{7}{*}{\rotatebox[origin=c]{90}{GloVe}} & STL & 364,042 & 1,785,967 &  364,042 & 1,785,967 \\
\cline{2-6}
     & NEG$_{CD}$ & 385,851 & 2,403,876 & 385,122 & 2,403,147 \\
\cline{2-6}
     & NEG$_{SFU}$ & 385,851 & 7,066,176 & 385,122 & 7,065,447 \\
\cline{2-6}
     & SPEC & 385,851 & 7,066,176 & 385,122 & 7,065,447 \\
\cline{2-6}
     & UPOS & 388,413 & 2,952,738 & 385,122 & 2,949,447 \\
\cline{2-6}
     & DR & 397,213 & 2,961,538 & 385,122 & 2,949,447 \\
\cline{2-6}
     & LEX & 1,191,204 & 3,755,529 & 385,122 & 2,949,447 \\
\hline
    \multirow{7}{*}{\rotatebox[origin=c]{90}{TL}} & STL & 1,001,170 & 56,870,931 & 1,001,170 & 56,870,931 \\
     \cline{2-6}
     & NEG$_{CD}$ & 1,051,939 & 56,921,700 & 1,051,210 & 56,920,971 \\
     \cline{2-6}
     & NEG$_{SFU}$ & 1,051,939 & 56,921,700 & 1,051,210 & 56,920,971 \\
     \cline{2-6}
     & SPEC & 1,051,939 & 56,921,700 & 1,051,210 & 56,920,971 \\
     \cline{2-6}
     & UPOS & 1,054,501 & 56,924,262 & 1,051,210 & 56,920,971 \\
     \cline{2-6}
     & DR & 1,063,301 & 56,933,062 & 1,051,210 & 56,920,971 \\
     \cline{2-6}
     & LEX & 1,857,292 & 57,727,053 & 1,051,210 & 56,920,971 \\
\hline
\end{tabular}
    \caption{Number of parameters for each model using different embeddings ordered by number of trainable parameters. The number of parameters is different for the MTL models depending on whether the parameters from the auxiliary task are included or not. The auxiliary task specific layer is shown as the pink layer in Figure \ref{fig:architecture}. The number of parameters including and not including the auxiliary task is stated as the MTL models at inference time would not use the auxiliary task parameters. There are many more trainable parameters for the MTL models ignoring the auxiliary task parameters. This is because the hyperparameter search finds a larger shared LSTM hidden dimension to be preferable for the MTL models (see Tables \ref{table:Parameters_STL} and \ref{table:Parameters_MTL}). For the GloVe MTL models the total number of parameters changes  depending on the auxiliary task. This is because the GloVe embeddings contain different numbers of vocabulary words, as we filter words based on those in the auxiliary and main task datasets/corpora. The large difference in the number of trainable parameters between GloVe and TL models is due to the fact that the TL is 724 parameters larger than the 300 parameter GloVe embeddings. Lastly, the number of trainable parameters is dataset agnostic, the number of all parameters is not dataset agnostic for the GloVe models due to the vocabulary size, for clarification the model parameters reported here are for those trained on the Laptop dataset.}
    \label{table:parameter_table}
\end{table*}

\begin{table*}[!h]
    \centering
    \begin{tabular}{|c|c|c|c|c|c|}
\toprule
Embedding & Model & Device  & Batch Size &   Min Time (s) &   Max Time (s) \\
\hline
     \multirow{16}{*}{GloVe} &   \multirow{8}{*}{STL} &           \multirow{4}{*}{CPU} & 1 &  10.24 &  10.45 \\
      \cline{4-6}
      &    &           &    8 &   7.00 &   7.21 \\
      \cline{4-6}
      &    &          &   16  &   6.67 &   6.91 \\
      \cline{4-6}
      &    &          & 32    &   6.35 &   6.51 \\
      \cline{3-6}
      &    &           \multirow{4}{*}{GPU} & 1 &   9.24 &   9.26 \\
      \cline{4-6}
      &    &           &    8 &   6.58 &   6.67 \\
      \cline{4-6}
      &    &          &   16  &   6.34 &   6.36 \\
      \cline{4-6}
      &    &          & 32    &   6.12 &   6.26 \\
      \cline{2-6}
      &   \multirow{8}{*}{MTL} &              \multirow{4}{*}{CPU} & 1 &  10.06 &  10.26 \\
      \cline{4-6}
      &    &           &    8 &   7.05 &   7.19 \\
      \cline{4-6}
      &    &          &   16  &   6.90 &   6.99 \\
      \cline{4-6}
      &    &          & 32    &   6.41 &   6.46 \\
      \cline{3-6}
      &    &           \multirow{4}{*}{GPU} & 1 &   9.43&   9.49 \\
      \cline{4-6}
      &    &           &   8  &   6.60 &   6.70 \\
      \cline{4-6}
      &    &          &   16  &   6.26 &   6.55 \\
      \cline{4-6}
      &    &          & 32    &   6.10 &   6.20 \\
      \hline
       \multirow{16}{*}{TL} &   \multirow{8}{*}{STL} &           \multirow{4}{*}{CPU} & 1 &  64.79 &  71.26 \\
       \cline{4-6}
        &    &            & 8    &  43.62 &  49.70 \\
        \cline{4-6}
        &    &           & 16    &  47.06 &  48.41 \\
        \cline{4-6}
        &    &           & 32    &  56.76 &  62.77 \\
        \cline{3-6}
        &    &           \multirow{4}{*}{GPU} & 1 &  23.26 &  23.79 \\
        \cline{4-6}
        &    &           &   8  &   8.82 &   9.09 \\
        \cline{4-6}
        &    &           &   16  &   8.57 &   8.86 \\
        \cline{4-6}
        &    &           & 32    &   8.45 &   9.78 \\
        \cline{2-6}
        &   \multirow{8}{*}{MTL} &               \multirow{4}{*}{CPU} & 1 &  64.01 &  67.90 \\
        \cline{4-6}
        &    &           &   8  &  49.05 &  50.00 \\
        \cline{4-6}
        &    &          &   16  &  53.47 &  56.42 \\
        \cline{4-6}
        &    &          & 32    &  55.33 &  55.79 \\
        \cline{3-6}
        &    &           \multirow{4}{*}{GPU} & 1 &  23.81 &  23.97 \\
        \cline{4-6}
        &    &           &   8  &   9.19 &   9.49 \\
        \cline{4-6}
        &    &          &   16  &   8.54 &   8.92 \\
        \cline{4-6}
        &    &          & 32    &   8.43 &   8.70 \\
\bottomrule
\end{tabular}
    \caption{Run/inference times for STL and MTL models that have been trained on the Laptop dataset using either GloVe or TL embeddings. Each model was timed in seconds (s) to generate predictions for 800 sentences, that were taken from the Laptop test split, of which this process was repeated five times and here we report the minimum (min) and maximum (max) time to generate predictions for those 800 sentences. We report these timings across different model configurations based on different batch sizes at prediction time and different devices. The trained MTL model used in this experiment was the MTL (NEG$_{SFU}$) version, this was chosen as it contains the largest number of total parameters as shown in Table \ref{table:parameter_table}. Further all of these times were based on the model already loaded into memory and using the Python timeit library for timings. Additionally the GPU used was a GeForce GTX 1060 6GB GPU, CPU was an AMD Ryzen 5 1600 CPU, and the computer had 16GB of RAM.}
    \label{table:inference_times}
\end{table*}

\end{document}